\documentclass{article}

\usepackage{arxiv}

\usepackage[utf8]{inputenc} \usepackage{ifthen}
\usepackage{csquotes}
\usepackage[english]{babel}
\usepackage[T1]{fontenc}    \usepackage[usenames,dvipsnames]{xcolor}
\definecolor{links}{rgb}{0.274,0.509,0.705}
\usepackage[colorlinks=true,allcolors=links]{hyperref}       \usepackage{url}            \usepackage{booktabs}       \usepackage{amsfonts}       \usepackage{nicefrac}       \usepackage{microtype}      \usepackage{graphicx}
\usepackage{amsmath}

\usepackage{setspace}
\usepackage[font=scriptsize]{caption}

\usepackage[backend=biber,
            style=apa]{biblatex}

\bibliography{benureau.bib}
\usepackage{marginnote}

\newcommand{\beginsupplement}{\setcounter{table}{0}
        \renewcommand{\thetable}{S\arabic{table}}\setcounter{figure}{0}
        \renewcommand{\thefigure}{S\arabic{figure}}}

\newcommand{\added}[1]{#1}
\newcommand{\removed}[1]{}

\title{Morphological Development at the Evolutionary Timescale:\\
       Robotic Developmental Evolution}

\author{
  Fabien C. Y. Benureau \\
  Okinawa Institute of Science and Technology, Japan\\
\texttt{fabien.benureau@oist.jp} \\
  ORCID: \href{https://orcid.org/0000-0003-4083-4512}{0000-0003-4083-4512}\\
  (corresponding author,\\
   permanent email: fabien@benureau.com) \\
\And
  Jun Tani \\
  Okinawa Institute of Science and Technology, Japan\\
\texttt{jun.tani@oist.jp} \\
  ORCID: \href{https://orcid.org/0000-0002-9131-9206}{0000-0002-9131-9206}\\
}

\begin{document}
\maketitle

\begin{abstract}
Evolution and development operate at different timescales; generations for the one, a lifetime for the other. These two processes, the basis of much of life on earth, interact in many non-trivial ways, but their temporal hierarchy—evolution overarching development—is observed for \added{most} multicellular lifeforms. When designing robots however, this tenet lifts: it becomes—however natural—a design choice. We propose to inverse this temporal hierarchy and design a developmental process happening at the phylogenetic timescale. Over a classic evolutionary search aimed at finding good gaits for tentacle 2D robots, we add a developmental process over the robots' morphologies. \added{Within a} generation, the morphology of the robots does not change. But from one generation to the next, the morphology develops. Much like we become bigger, stronger, and heavier as we age, our robots are bigger, stronger and heavier with each passing generation. Our robots start with baby morphologies, and a few thousand generations later, end-up with adult ones. We show that this produces better and qualitatively different gaits than an evolutionary search with only adult robots, and that it prevents premature convergence by fostering exploration. In addition, we validate our method on voxel lattice 3D robots from the literature and compare it to a recent evolutionary developmental approach. Our method is conceptually simple, and can be effective on small or large populations of robots, and intrinsic to the robot and its morphology, \added{not the task or environment.}\removed{, and thus \added{hopefully} not specific to the task and the fitness function it is evaluated on.} Furthermore, by recasting the evolutionary search as a learning process, these results can be viewed in the context of developmental learning robotics.
\end{abstract}

\keywords{Development \and Evolution \and Learning \and Robotics \and Evolutionary Computation}
\newpage

\section{Introduction}

Generations evolve; individuals develop. Many of the evolutionary processes of a species, then, happen at a timescale an order of magnitude or more greater than the developmental ones: evolution is a long-term species-level process happening on top of many short-term individual developmental processes. This is the rhythm of our biological world.

Robots do not have to abide by such principles. Here, we propose to inverse the timescales, and put a developmental process on top of an evolutionary one. We consider an evolutionary process with robots; the first-generation robots are small and weak and light. They are not the target morphology: only baby versions of it. They attempt to solve a task; the best performers survive and produce slightly mutated offspring of themselves for the next generation. Those second-generation offspring are a bit bigger, a bit stronger, and a bit heavier than the first generation. The second generation, in essence, \emph{grows up} compared to the first one. Following a predefined developmental schedule happening at the phylogenetic timescale, each generation will be bigger and stronger and heavier than the previous one until reaching adult morphology. Our robots do not develop during fitness evaluation: their morphology is determined at birth and remains constant while they are evaluated on the task. It is a developmental process \emph{for generations}, rather than for an individual. Instead of \emph{evolutionary developmental robotics}, this is \emph{developmental evolutionary robotics}: \emph{devo-evo-robo} rather than \emph{evo-devo-robo}.

To understand some of the motivations behind this work, another perspective on it is useful. Above, we are proposing a new method for evolutionary robotics, adding development on top of evolution, and demonstrating that it can improve which behaviors are discovered  and trying to analyze why it does. Yet, another view is to consider the evolutionary search as a rudimentary trial-and-error learning process employed by a single robot: in that context, a generation of 20 robots becomes an epoch of 20 trials, and the 4000 generations of the evolutionary search turn into our lone robot's lifetime: 4000 epochs of learning. \emph{Developmental evolution} becomes \emph{developmental learning} \parencite{Lungarella2003, Cangelosi2015}, and our initial question: "Can development improve the evolutionary process?" becomes "Can growing up help a robot learn better?" which may help, in turn, \added{to} shed light on the question: "Does growing up help \emph{us} learn better?". Morphological computation \parencite{Pfeifer2009} looks at how morphology participates in behavior. We are interested here in how morphology, or more precisely, morphological change, participates in behavior acquisition. Those underlying questions critically motivate our investigation.

The overwhelming majority of animals go through morphological growth at the beginning of their lives, yet the research on growing robots is scarce, and most of them mimic the indeterminate growth of plants \parencite{DelDottore2018, Corucci2017}. Most of the previous work on development in evolutionary robotics has focused on adding developmental processes to the fitness evaluation phase, i.e. \emph{evo-devo} approaches \parencite{Kriegman2017,Kriegman2018,Kriegman2018b,Bongard2011,Corucci2016}. While this mimics biology at the phylogenetics timescale, computational and time constraints limit the extent and complexity of the developmental program, which must fit entirely in an evaluation period, usually a few dozens of simulation-seconds long. \parencite{Vujovic2017} illustrates this: by considering real-world robot development within a generation, and having to construct and evaluate three different robot morphologies for each fitness evaluation, the number of generations, five, remains drastically limited. Furthermore, because even in simulation, fitness evaluations rarely exceeds a few hundreds of seconds, this also forces development to happen at the same timescale as behavior, which may be problematic if modeling biology. By contrast, development slowly rolling out over generations has the potential to implement complex developmental paths, without adding any more computational cost per generation than a classical evolutionary algorithm, \added{while allowing development to happen at an arbitrary slower timescale than a single behavior} \removed{and behavior and development happen separately}. To our knowledge, only Josh Bongard explored a \emph{devo-evo} approach, studying a robot going through four different morphologies over the course of evolution, from anguilliform to legged adult, to conclude that it made the task more difficult, worsening performance, and proposing as a better-performing approach one where robots develop over the evaluation period \parencite{Bongard2011}.

Our approach is an incremental evolution one \parencite{Mouret2008, Doncieux2014}: the evolutionary search starts with an altered version of the target task, and then, incrementally as the search progresses, is transformed into the target task. There are several methods in incremental evolution: staged evolution (splitting a difficult task into subtasks) \parencite{Parker2001, Mouret2006, Urzelai1998, Urzelai1999}, behavior decomposition (training sub-controllers), fitness shaping (modifying the fitness function to create smoother gradients) \parencite{Nolfi1997, Colby2015}, and environment complexification (mak\added{ing} the environment/task harder or more complex as the search progresses) \parencite{Gomez1997, Bongard2011b, Miras2019, Miras2019b}. Our approach falls into the environment complexification methods, although \added{that, in contrast to most methods}, we modify the robot itself—its morphology—rather than the rest of the environment.

Our approach distinguishes itself from existing incremental methods in two ways. First, most incremental approaches work by making the problem easier and working their way up to the target task: splitting a complex task in simpler subtasks, training subcontrollers, making the fitness function less rugged or less sparse, and progressively complexifying the environment---for instance progressively increasing a prey speed when training a predator behavior \parencite{Gomez1997}. By contrast, our approach is not to make the problem obviously easier. Indeed, much like human babies do not have an easier time to walk than adults, our robots start with immature bodies that are less apt are walking than adults. We would expect a decrease in \added{final, adult} performance\added{, as a significant part of learning happens on a different, and in some cases obviously inferior morphology}. This is actually one of the main theoretical contribution of this article: despite starting with immature bodies, our population of robots discover better behaviors. Morphological development here, rather than being a hindrance, helps.

Second, incremental evolution approaches have been shown to be effective, but they necessitate to design a tailored incremental program to modify the task or environment. This requirement for hand-tuned expert knowledge on a case-by-case basis severely limits their generality and introduces bias into the solution. Environmental and task complexification mirrors teaching in humans, where the best results are obtained by an attentive teacher tailoring a curriculum of lessons and exercises of increasing difficulty to the student, with a change of topics requiring to design a new curriculum, usually from scratch. By contrast, the fundamental mechanisms of human development have changed little over the last thousands of years, while the environment, the tasks we engage in and the skills we develop have undergone dramatic transformations. Morphological development in robots is modelled after the latter. It does not require any modification of the environment or the task; it is intrinsic to the robot itself and to its embodiment. This opens the possibility that there exists, much like in humans, developmental programs for a given robot that can be effective in a wide range of task and environments. While this possibility is a major motivation for our study, this is not something we establish here, and for the time being, we have to rely on hand-tunned developmental paths.

But there's an even stronger argument to be made there: as we observe that humans, after an extended developmental period, end up possessing generalization capabilities unmatched across the animal kingdom \parencite{Smith2005, Anderson2003, Shapiro2011}, we may formulate the hypothesis that development is not just able to adapt, but also \emph{needed} to adapt to a range of different tasks, needed for the generalization capabilities of humans. Morphological development may one day participate in making robots, as Linda B. Smith says, \emph{flexibly smart} \parencite{Smith2005}. It goes without saying that our results are limited and simplistic in most of their aspects, and prove little robotically and nothing biologically. Still, those questions and hypotheses underpin the work presented.

The developmental trajectories we explore in this article are inspired by morphological growth in the animal world: change in size, in mass, in muscle strength. In particular, we consider a simplified model of muscle development for our robots, where those three dimensions develop together under physical constraints. With this simple physically-plausible muscle model, we show that robot generations that grow up discover better ways to move than robot generations that only feature adult robots. Then, we study how development affects exploration within the evolutionary process.
 \section{Method}

We consider simple "starfish" soft-robots in a 2D world. The robots possess a hexagonal body, and six tentacles stem from it. The robots are subjected to gravity and laid on a flat floor. They are evaluated on their ability to move as far as possible along the floor in 60 seconds. An evolutionary search is performed over the gaits of the robots.

\subsection{Robots}

We consider a physics engine with only three basic elements: point masses, springs between point masses, and a flat floor that collides with the point masses. Each point mass experiences reaction forces and friction forces from the ground, gravity, as well as the forces of the springs connected to it. Given a spring linking two point masses of mass $m_A$ and $m_B$, with resting length $x_r$, length $x$, stiffness $k$ and damping ratio $\zeta$, we have:
\[m_r\ddot{x} = - k (x-x_r) - c \dot{x} \quad\textrm{with}\quad m_r = \frac{1}{\frac{1}{m_A} + \frac{1}{m_B}} \quad\textrm{and}\quad c = 2\zeta\sqrt{m_r k}\]

Our robots have passive springs for which the resting length is fixed, and actuated springs, where the resting length is modified by the motor commands of the robot: decreased for contraction, and increased for extension. We will refer, in the rest of the text, to actuated springs as the artificial \emph{muscles} of the robot.

\begin{figure}[t]
  \centering
  \includegraphics[width=16cm]{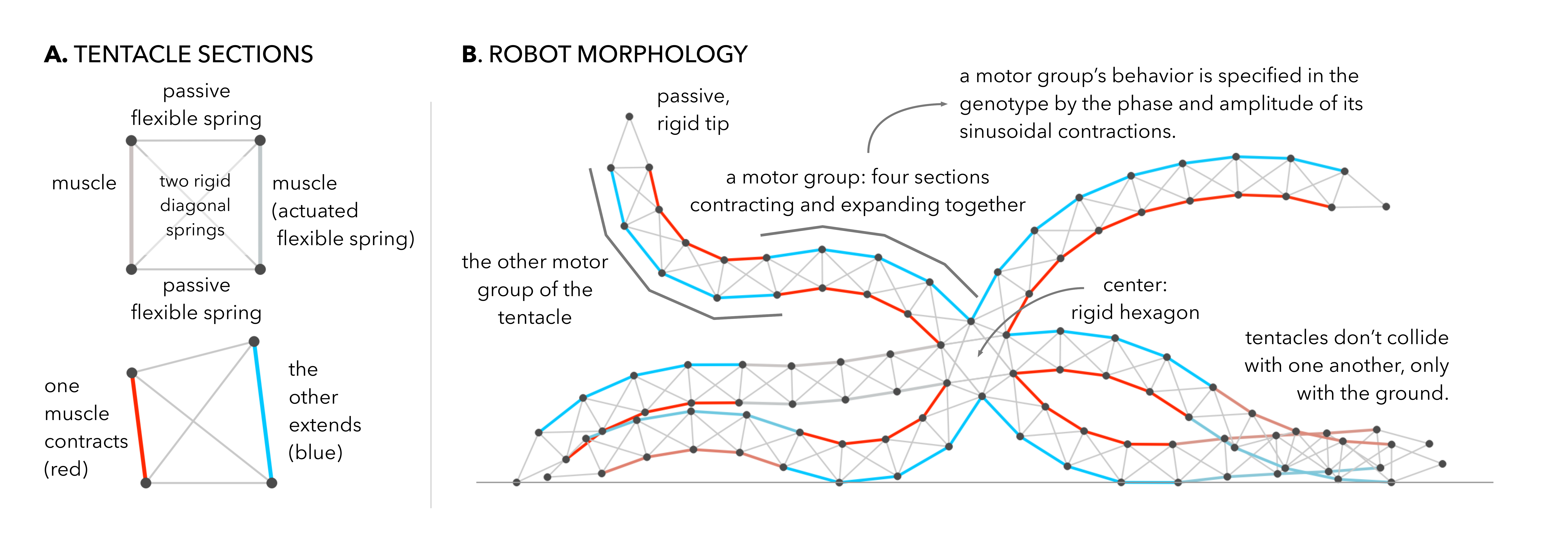}
  \captionsetup{width=16cm}
  \caption{\textbf{Our robots are made of point masses and springs.} \textbf{A.} A section is composed of two passive and rigid (high-stiffness) diagonal springs, two passive and flexible springs on the top and bottom and two actuated springs, the \emph{muscles}, on each side. The muscles work in pairs and act in an antagonistic manner: when one contracts (by decreasing its rest length, in red), the other extends (in blue). The color represents \added{the command sent to the muscle, i.e.} the desired length of the muscle, not the actual length. \textbf{B.} The sections are assembled into tentacles, with a passive triangular tip at the end. Sections are grouped into motor groups that actuate together, and attached to a central hexagonal body.}
  \label{fig:section}
\end{figure}

Our robots follow a starfish pattern: a main body—a regular hexagon made of rigid springs—from which six tentacles stem (Figure \ref{fig:section}). The tentacles are composed of eight sections, plus a passive triangular tip. Each section is a square with rigid diagonal springs, and passive flexible springs at the top and bottom. On each side of the section, which also are the side of the tentacle, a muscle---an actuated spring---is present. All the nodes of a given robot have the same mass, and all its muscles have the same stiffness. The stiffness of passive springs is the same in all robots (see supplementary data for a detailed description).

The two actuated side-springs work as antagonistic muscles: when the tentacle section receives a motor command, one muscle contracts while the other muscle extends. The motor command of a section is a scalar, received at each timestep; if a section of height $h$ receives the motor command $\alpha$, the target length—i.e., the rest length of the spring—of the left muscle becomes  $(1 + \alpha) h$, while the target length of the right muscle becomes $(1 - \alpha) h$.

Sections are gathered into motor groups (Figure \ref{fig:section}.B). An eight-section tentacle is divided into two groups of four sections each. All the sections of the same group receive the same motor command, and therefore expand and contract simultaneously.

The gait of the starfish applies a sinusoidal actuation signal over each motor group. The period of the sinusoid is fixed (at $2\pi$) and shared by all the motor groups, and each motor group has an independent phase (in $[-\pi, \pi]$) and amplitude (in $[0, 0.2]$). This allows to encode the gait of the starfish with two scalars per muscle group, and thus with 24 scalars in total for the starfish: those 24 scalars form the genotype of our robots.

\subsection{Task}

The robots must move to the right as far a possible in 60 seconds; the fitness is the distance covered. The robots are dropped just above the ground. To avoid the robots taking advantage of the drop to bounce off the ground, which may produce chaotic fitness values, the robot settles for 9.42 ($3\pi$) seconds on the ground with no motor activation (0.0 actuation value sent to all motor groups). The robot then actuates until the 60-second mark, resulting in eight full actuation periods.

We employ a simple evolutionary strategy. At each generation, the five members with the best fitness become parents of the next generation, each creating three children through mutation. The parents survive too, creating a new population composed of the five parents and the 15 children.

Typically, in evolutionary robotics, the search is conducted from start to finish on the target---adult---morphology. We will call this \emph{adult evolution}, and it will serve as control to evaluate the performance of \emph{developmental evolution}, which we explain now.

\subsection{Developmental Evolution}

\begin{figure}[ht]
  \centering
  \includegraphics[width=15cm]{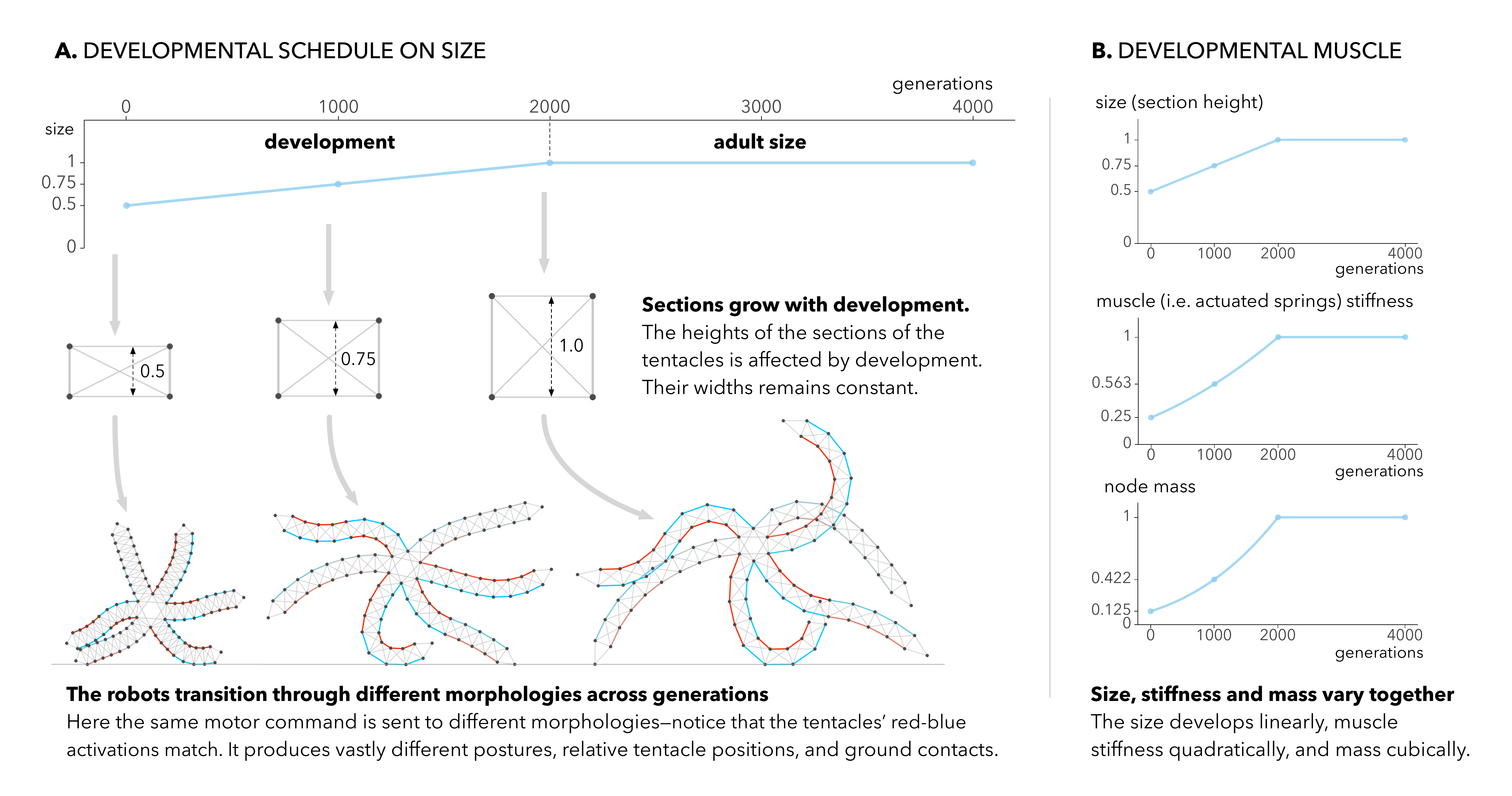}
  \captionsetup{width=15cm}
  \caption{\textbf{The robots grow up according to a predefined schedule.} \textbf{A.} From generation 1 to 2000, the height of the tentacle sections grows linearly from 0.5 in the first generation to 1.0, the adult size. It remains there for generation 2000 to 4000. \textbf{B.} The developmental muscle model combines the development of three morphological characteristics of the muscle in parallel: size, muscle strength and node mass.}
  \label{fig:size}
\end{figure}

Developmental evolution considers populations of robots that develop across generations. During fitness evaluation, the morphology of the robots remains fixed and is the same for all members of the generation. But, from one generation to the next, the morphology changes slightly. This change is not under evolutionary control. It is predefined and goes according to a fixed schedule.

Figure \ref{fig:size}.A shows an example. In the first generation, the robots start at size 0.5: the height of the tentacles sections (including the tip), is half the height of the adult morphology. Each subsequent generation will see the size increment linearly (by 0.5/2000 = 0.00025) until generation 2000, when the adult morphology (size 1.0) is reached. For the next 2000 generations, until generation 4000---the end of the search---the robots will keep this adult morphology. In our evolutionary algorithm, the parents survive to the next generation. Therefore, their morphologies are adjusted to the new developmental values and they undergo fitness evaluation again.

Size is not the only morphological characteristic whose development we study. We analyze how developing muscle strength and the mass of the nodes of the robot affect the evolutionary process, including what happens when those three variables---size, strength, mass---develop together. \removed{This, we do in the} \added{For this, we create a crude} developmental muscle model.

\subsection{Developmental Muscle Model}

In animals, muscle strength is a function of an array of factors. One of the most prominent ones is the cross-sectional area of the muscle, which has a proportional effect on strength \parencite{Sacks1982}. If a muscle doubles in size in every proportion, its cross-sectional area quadruples, and therefore the strength is affected in a quadratic fashion. Meanwhile, the mass of the muscle, dependent on volume, is affected cubically.

We use these insights to create a simple developmental muscle model for our robots that combines the development of size, strength, and mass. In our model, when the size doubles, the muscle strength—i.e. spring stiffness—quadruples and the node mass is multiplied by eight. We express size, stiffness and mass as coefficients of the adult values (see supplementary data for a description of the robots), and therefore the adult value for all characteristics is 1. For a given size coefficient $s$, the stiffness coefficient is $s^2$ and the mass coefficient $s^3$ (Figure \ref{fig:size}.B).
 \begin{figure}[htbp!]
  \centering
  \includegraphics[width=14cm]{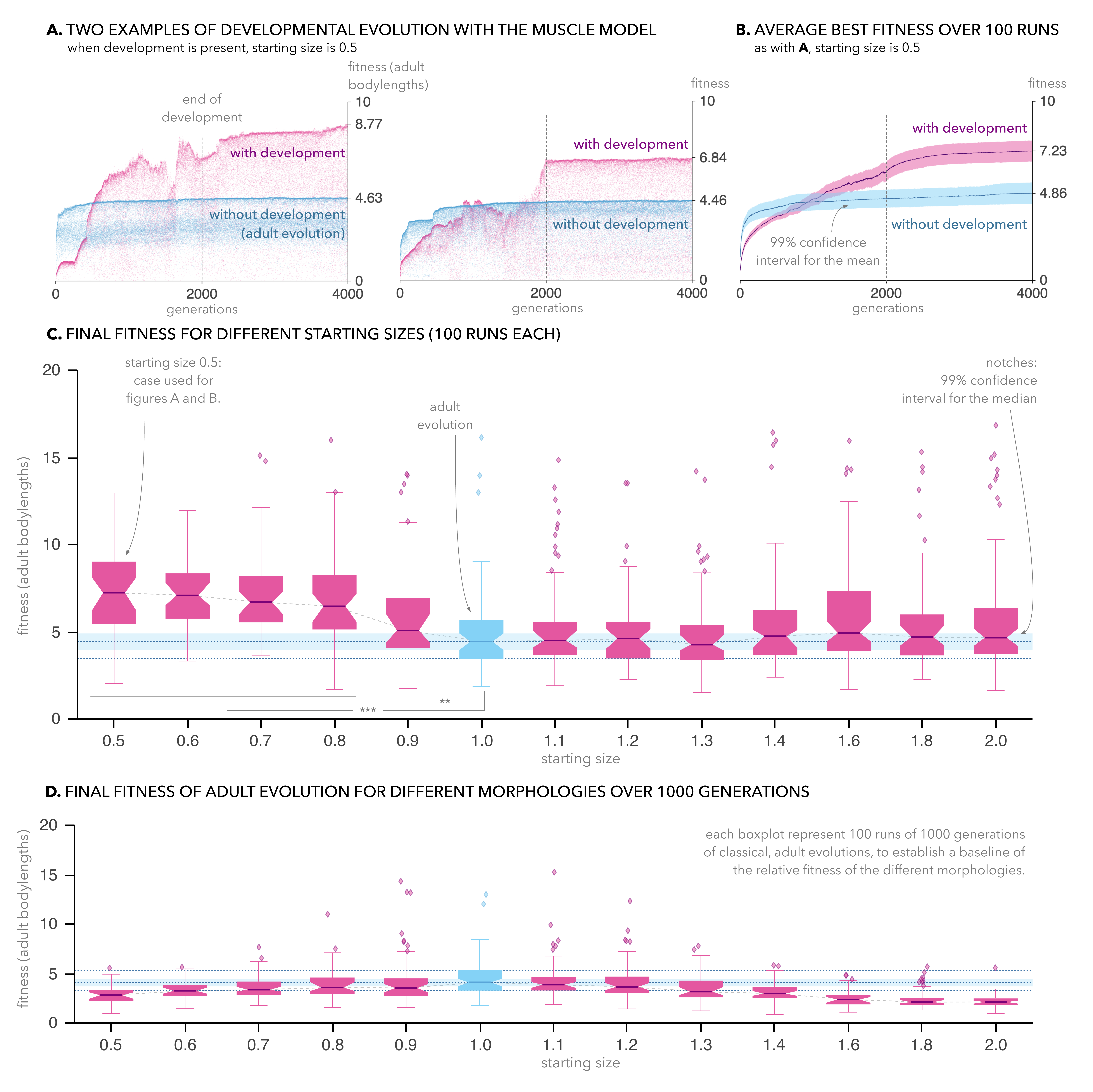}
  \captionsetup{width=14cm}
  \caption{\textbf{The developmental muscle model outperforms adult evolution.} \textbf{A.} \removed{Two examples of developmental muscle model developmental evolution with starting size 0.5}\added{Two evolutionary runs of developmental evolution with the muscle model at starting size 0.5}, out of the 100 runs performed, contrasted with the adult evolution performed with the same initial population. Each dot represents the fitness of a population member. Not shown: members with negative fitness. \textbf{B.} Average \removed{best}\added{final} fitness over the 100 runs of the developmental muscle model with starting size 0.5 and the adult evolution. \removed{The best fitness is the highest fitness for each generation of each run}\added{The final fitness is the mean of the ten best fitnesses obtained in the last 50 generations}. The shaded area is the 99\% confidence interval of the mean \added{of the final fitness over 100 runs}. \textbf{C.} Comparison of the distribution of the final fitness of the developmental muscle model for different starting sizes. \removed{The final fitness is the mean of the ten best fitness obtained in the last 50 generations.} Boxplots are computed from 100 runs each, and the same set of 100 random seeds is used for each boxplot, creating paired experiments with the same initial populations. Boxplots show the first and third quantile, and the minimum and maximum value, or $1.5 \times \textrm{IQR}$, whichever is closer to the median; outliers are represented by diamonds. Notches represent the 99\% confidence interval of the median. Significance stars are computed using a Wilcoxon signed-rank test after Shapiro-Wilk testing revealed that the differences between some of the paired fitnesses deviate significantly from normality. The significance threshold is set at 0.01. These conventions will be used for all boxplots of the article. In blue, the adult evolution, equivalent to a starting size of 1.0, has its mean, confidence interval, and first and third quantile extended through the plot by a light shaded blue area and dashed horizontal lines. \textbf{D.} Adult evolution over 1000 generations over all the starting morphologies of C. No development is involved.}
  \label{fig:dev_muscle_results}
\end{figure}

\section{Experiments \& Results}

\subsection{Developmental Muscle Model Experiment}
\label{sec:results_dev_muscle}

We run evolutionary searches with the developmental muscle model and contrast them with the adult evolution, where the robots are adult from the first generation. In Figure \ref{fig:dev_muscle_results}.A, two \removed{examples}\added{evolutionary runs} of developmental evolution using the developmental muscle model are shown, where the robots start with a size of 0.5 in the first generation, and with the corresponding values of 0.25 for muscle stiffness and 0.125 for the mass of the nodes (see \hyperref[sec:movies]{Movie S1} for behavior throughout the generations for the first graph.). They then gradually increment toward the adult values for the size, stiffness, and mass respectively for the next 2000 generations. In the two examples, the adult evolution displays an archetypical learning curve, which progresses quickly at the beginning and then flattens. In contrast, the muscle model's developmental phase is characterized by sudden changes in fitness, both increases and collapses. Some of the behaviors, especially in the first graph, have higher fitness on young morphologies than the one found by the adult evolution. After the development ends, the fitness stabilizes around a significantly higher value than the adult evolution.

This difference in performance is statistically confirmed by Figure \ref{fig:dev_muscle_results}.B, where the average fitness over generations for 100 repetitions of the experiment is shown (for a given condition, the 100 repetitions use the random seeds 1, 2, ..., 100 and therefore the same initial population of genotypes, creating paired experiments across different conditions). We observe a steady rise of fitness during development, with the average over 100 repetitions smoothing out the sudden collapses and increases \added{of an individual run}. At the end of development, the fitness is already significantly better than the adult evolution (p~<~0.001, Wilcoxon signed-rank test). The average fitness then continues to increase after generation 2000 for a few hundred generations at a larger rate than the adult evolution before stabilizing.

\begin{figure}[thbp!]
  \centering
  \includegraphics[width=14cm]{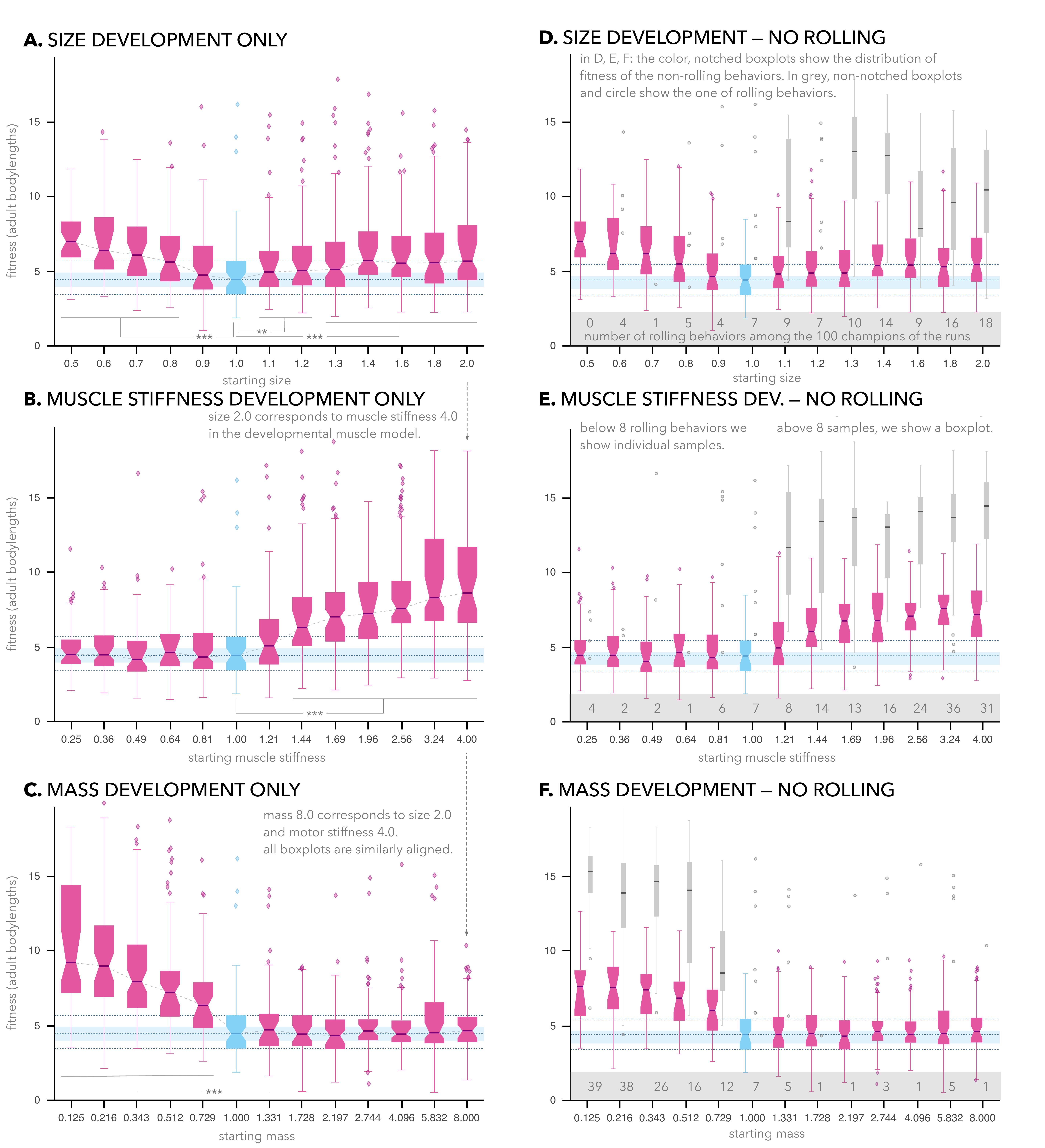}
  \captionsetup{width=13cm}
  \caption{\textbf{Size, muscle stiffness and mass development can all benefit the evolutionary process.}  The experiments in \textbf{A.}, \textbf{B.}, and \textbf{C.} show the effect of the development that happened in Figure \ref{fig:dev_muscle_results}.C when size, muscle stiffness and mass develop on their own respectively, while the two other characteristics are fixed at adult values. Boxplots conventions are the same as in Figure \ref{fig:dev_muscle_results}. \textbf{D.}, \textbf{E.}, and \textbf{F.} show the same data as \textbf{A.}, \textbf{B.}, and \textbf{C.}, respectively, with the rolling (in grey) and non-rolling behaviors (in color) separated. The non-rolling behaviors are represented by boxplots when there are 8 samples or more, and otherwise by grey circles (also used for outliers). Due to the small and varying sample size, the rolling boxplots are not notched.}
  \label{fig:sizestiffmass_boxplots}
\end{figure}

Figure \ref{fig:dev_muscle_results}.C gives insights into the effect of different developmental trajectories by comparing different starting sizes for the developmental muscle model. Starting smaller, weaker, and lighter provides a sizeable and significant increase in performance compared to the adult evolution, with a peak observed for starting size 0.5. This increase is reduced when the starting values of the development get closer to the adult ones, but even a modest development—starting size 0.9—provides notable benefits to fitness. Starting bigger, stronger, and heavier is not as effective, and does not bring significant benefits.

Interestingly, none of the starting morphologies with the developmental model is better at solving the task than the adult morphology, as Figure \ref{fig:dev_muscle_results}.D shows. Over 1000 generations, all the starting morphologies are learned on a standard adult evolution, to establish a baseline of the fitness they can reach. The results show that development can provide higher fitness by passing through morphologies that are worst at the task than the target one. Moreover, given that the fitness is already better at generation 1999 in Figure \ref{fig:dev_muscle_results}.B, developmental evolution can provide better fitness on the adult morphology without ever directly experiencing it.

To understand the causes behind those dynamics, it is useful to look at the effect of developing the size, the muscle stiffness, and the mass separately. In each instance, the two other characteristics are fixed at adult values during the entire evolutionary process. The results are shown in Figure \ref{fig:sizestiffmass_boxplots}.

Perhaps unsurprisingly, having low mass or being strong at the beginning brings unequivocal performance increases. These results would be easily explainable if those characteristics---low mass or high stiffness---were permanent. But the performances displayed here are the ones of the adult morphology, after development ended\added{, since they are calculated from the mean of the 10 best individuals of generations 3951--4000}. It is interesting to notice that the increased fitness of the low-mass or high-stiffness start have increased variability across different runs. What's hidden here is that we have two populations of behaviors: rolling and non-rolling behaviors. For our purposes here, we will define a rolling behavior as the robot's central body having done two or more complete revolutions on itself by the end of the evaluation. Typically, non-rolling behavior manifests as crawling or shuffling  (see \hyperref[sec:movies]{Movie S2} and \hyperref[sec:movies]{S3} for example of non-rolling and rolling behavior respectively). Rolling is a highly beneficial behavior and allows to reach fitness scores that non-rolling behavior does not. This is shown with Figures \ref{fig:sizestiffmass_boxplots}.D, E, and F: the distributions of fitness of the rolling (in \removed{color}\added{grey}) and non-rolling (in \removed{grey}\added{color}) behaviors are separated. We observe that the low-mass and high-fitness developments generate a lot of rolling behaviors compared to adult evolution, which is hardly surprising: it is easier to start a roll when light or strong. The evolutionary process is then able to refine and adapt the rolling behavior to the changing morphology, retaining them into adulthood.

Conversely, when handicapping the robot during development by a high mass or low stiffness, no significant fitness benefit is observed, and the number of rolling behavior discovered is even lower than adult evolution: indeed, the robot has difficulties moving from its initial position when development starts, much less engage in a highly dynamical behavior such as rolling. It is notable though that these handicapping developmental paths do not have a long-term detrimental effect: the final fitness is not significantly different from the control. Development, here, is robust and degrades gracefully.

\added{Here,} it is interesting to remark that the difference between morphological development and environmental scaffolding is sometimes thin. In particular, developmental evolution with a low starting mass is quite similar---although not equivalent---to environmental scaffolding with a low starting gravity. Figure \ref{fig:s_gravity} confirms this; the performance impact of both interventions is remarkably similar. Of course, changing gravity outside of simulations is hardly ever a practical method.

Developing the size brings interesting results. Starting small generates the most benefits, and the benefits increase the smaller the robots start. Few rolling behaviors are discovered. \removed{However}\added{Remarkably}, when comparing the fitness of the non-rolling behaviors alone across experiments (Figure \ref{fig:sizestiffmass_boxplots}.D, E, and F), the benefits of development for starting small, or strong, or light are similar. Starting big is helpful as well, with more rolling behaviors discovered. One hypothesis to explain the increase in rolling behavior is that longer tentacles tend to break (see
Figure \ref{fig:tentacle_breaks}), and thus are not as much an obstacle to rolling as they might seem. Here the importance of \emph{discovering} the behavior seems paramount, and once discovered, maintenance of it by the evolutionary process through the morphological changes seems comparatively easier, even as tentacle breaks become less and less possible as the morphology nears the adult body.

Overall, we observe here two important dynamics. First, developmental evolution can generate better behaviors and help discover efficient behaviors that are hard to find if only experiencing the adult morphology. And second, different developmental paths lead to different behaviors: starting small is highly beneficial, it improves the behavior of the robots significantly, but it will not induce many rolling behaviors, while a high-stiffness or low-mass developmental path will.

It remains to be explained, though, why non-rolling behaviors are improved as well by developmental evolution. To do that, it is useful to look at how the method impacts the exploration of the search space.

\subsection{Development Fosters Exploration}

\begin{figure}[htpb!]
  \centering
  \includegraphics[width=13cm]{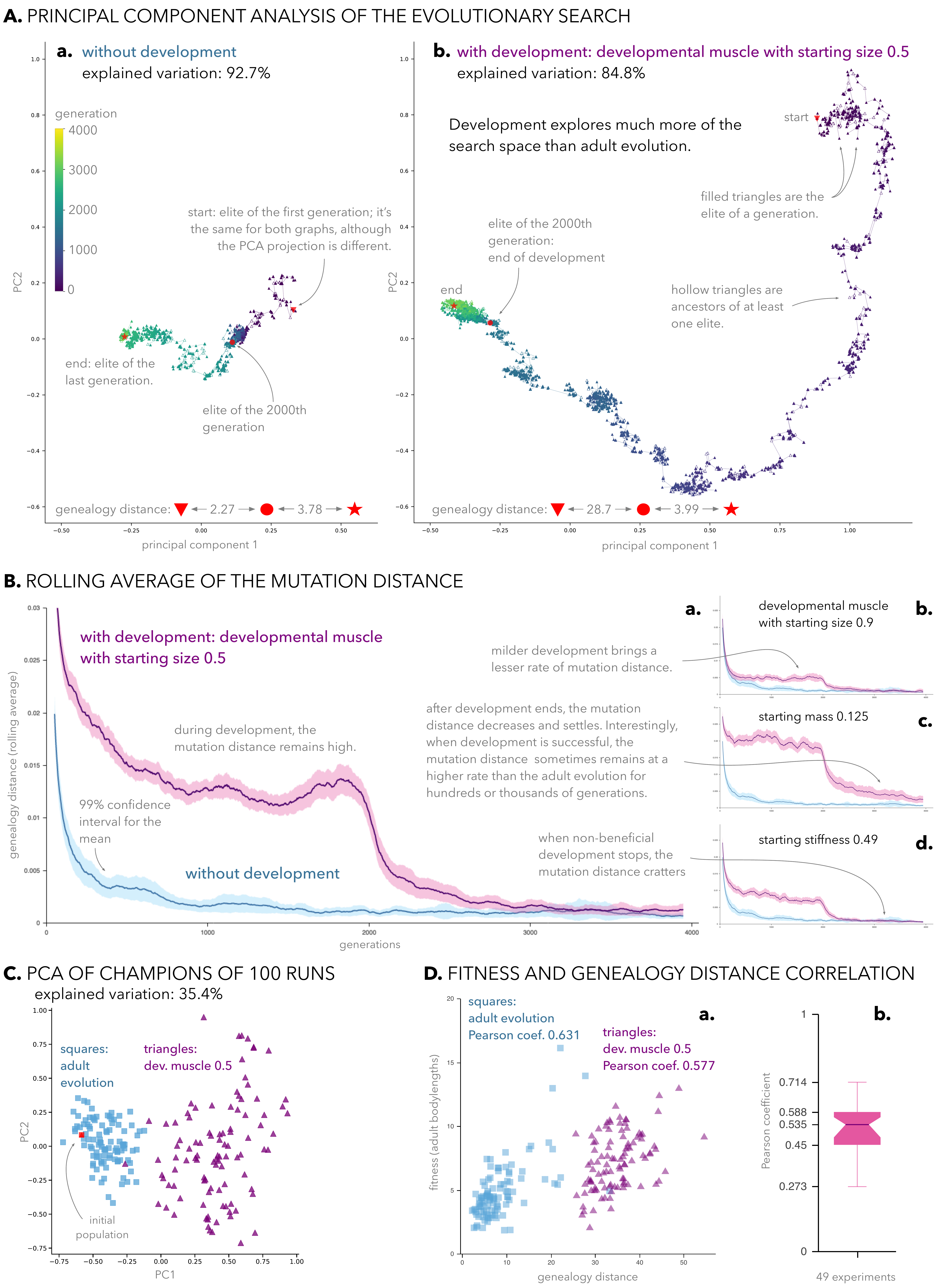}
  \captionsetup{width=13cm}
  \caption{\textbf{Development has a notable impact on exploration}. \textbf{A.} Principal component analysis of the genotype vectors of two different runs: one with the developmental muscle model with starting size 0.5 (\textbf{A.b}), and one with the adult evolution (\textbf{A.a}). The starting population was precomputed and the same in both cases, and was composed of only one member (red inverted triangle), selected as the one out of 15 random members with the best fitness on the adult morphology. The genealogy shows the member with the best fitness of each generation (filled triangles), alongside all their ancestors (unfilled triangles). Edges show parental relationships. The PCA was computed separately for each run, on all the depicted genotypes.
  \textbf{B.} Rolling average with a 101-generation window of the mutation distance of each generation, for four different developmental cases: two with the muscle model, one with mass development only, and one with stiffness development only.
  \textbf{C.} Given the same initial population, with only one member, PCA of the genotype of the champion of each run, for the adult evolution (squares) and the developmental muscle with starting size 0.5.
  \textbf{D.} Genealogy distance correlates positively with fitness. In \textbf{D.a}, the distribution of fitness in function of the genealogy distance of the elite at the end of the evolutionary search is displayed. The correlations are computed separately for control and the developmental muscle case with starting size 0.5. \textbf{D.b} Distribution of the fitness/genealogy distance correlation for all the experiments of Figure \ref{fig:dev_muscle_results}.C (adult evolution + 12 starting sizes) and \ref{fig:sizestiffmass_boxplots} (12 conditions for starting size, stiffness and mass), all with a significant correlation ($p < 0.01$).}
  \label{fig:gen_dist_pca}
\end{figure}

For the first half of the evolutionary search, developmental evolution changes the morphology of the robots between every generation. Therefore, the fitness landscape changes with each generation during that time. Gaits that were competitive on a morphology are not as much on the next one, and are quickly superseded by new gaits: development, quite straightforwardly, prevents convergence. This can be observed in Figure \ref{fig:gen_dist_pca}.A, which shows a principal component analysis of the champions of each generation (filled triangles), plus their ancestors (hollow triangles), for the adult evolution and the developmental muscle model with starting size 0.5. Developmental evolution travels much more than adult evolution across the search space, and is especially mobile during the developmental period.

To quantify it, let's define the \emph{genealogy distance} between a member of the population and one of its ancestors as the sum of all the \emph{mutation distances}—i.e., the euclidean distances—between successive members of the genealogy that goes from the ancestor to the member we are considering.
In Figure \ref{fig:gen_dist_pca}.A, the genealogy distance between the initial root member and the elite of the 2000th generation is 28.75 (over the 100 runs: 30.28$\pm$0.74, mean with 99\% confidence interval), versus 2.27 (6.66$\pm$0.84) for the adult evolution. From generation 2000th to the elite of the 4000th generation, the genealogy distance of the elites of generation 4000 is 3.99 (3.48$\pm$0.59) versus 3.78 (1.81$\pm$0.45) for the adult evolution.

This effect can be seen again in Figure \ref{fig:gen_dist_pca}.B, averaged over 100 repetitions of the experiments of \ref{fig:gen_dist_pca}.A. Given the champion of the run, the graph shows the rolling average of the mutation distance between each of the successive generations of the champion's ancestors. The rolling average of the mutation distance is high during the developmental period and drops as soon as development ends, and therefore as soon as the fitness landscape stops changing. This increase is correlated to the magnitude of development. The developmental muscle model with starting size 0.5 (\ref{fig:gen_dist_pca}.B.a) see a higher rise in mutation distance than the one with starting size 0.9 (\ref{fig:gen_dist_pca}.B.b). For some of the most successful developmental evolution, such as the one with starting mass 0.125 (\ref{fig:gen_dist_pca}.B.c), a significantly higher rate of mutation distance is maintained over the adult evolution for a long time after the stabilization into adult morphology. Conversely, non-beneficial development, such as the low starting stiffness 0.49 (\ref{fig:gen_dist_pca}.B.d), still displays high mutation distance during development but quickly drops to adult evolution levels when development ends.

Another way to illustrate this is via Figure \ref{fig:gen_dist_pca}.C: from the same initial genotype, 100 runs of the adult evolution and the developmental model with starting size 0.5 were computed. Looking at a PCA of the genotype of the champion of each run, the pattern is clear: the adult evolution clusters around the initial search point, while the developmental runs went far away from the initial population, and all in the same general direction.

While development is indeed a way to prevent (premature) convergence by pushing the evolutionary search to move around a constantly evolving fitness landscape, it would not be useful if the location it arrived at at the end of development, at generation 2000, was not any better than the random ones provided by the initial population\added{, or the learned ones by adult evolution by generation 2000}. The fitness results of section \ref{sec:results_dev_muscle} show that development can indeed be beneficial, but Figure \ref{fig:gen_dist_pca}.B.d \added{(starting stiffness 0.49)} and the corresponding performance in \ref{fig:sizestiffmass_boxplots}.B equally demonstrates that moving around during development does not guarantee better performance. Still, when looking systematically at the relationship between exploration and fitness, some patterns are present.

It turns out that genealogy distance and fitness are correlated \removed{even} \emph{within} a given developmental \removed{path}\added{condition}, as shown in Figure \ref{fig:gen_dist_pca}.D. Figure \ref{fig:gen_dist_pca}.D.a exemplifies the correlation that can be found between fitness and genealogy distance, whether for the adult evolution or the developmental muscle model. \removed{Figure \ref{fig:gen_dist_pca}.D.b, show the distribution, for all the experiments of the Figure \ref{fig:dev_muscle_results}.C and \ref{fig:sizestiffmass_boxplots}, of the correlation between the genealogy distance and the fitness of the champion over the 100 runs of each. The median Pearson correlation coefficient is 0.535.} Remark here that we do not compare developmental and adult evolution: the correlation is \emph{within} the 100 runs of a given condition. The runs that travel more throughout the search space for a given developmental schedule (or the adult evolution) \removed{also} tend to achieve higher fitness. \added{Figure \ref{fig:gen_dist_pca}.D.b explores how different amount of exploration of the search space from one developmental condition (i.e. one experiment) to the next correlate with higher fitness. Across all experiments of Figure \ref{fig:dev_muscle_results}.C and \ref{fig:sizestiffmass_boxplots} \added{(49 experiments of 100 runs each)}, we correlate the fitness of the best individual across 100 runs (chosen within the last 50 generations) to its genealogy distance to its ancestor in the initial population. The median Pearson correlation coefficient is 0.535: developmental conditions that travel more through the search space tend to generate better performance. This correlation is present within runs (Figure \ref{fig:gen_dist_pca}.D.a) and across experiments (Figure \ref{fig:gen_dist_pca}.D.b).}

These results do not provide a conclusive answer to why developmental evolution improves the effectiveness of the non-rolling behaviors, but it strongly suggests one: a better exploration of the search space.

\subsection{Population Size, Development Length, and Fast Development}

\begin{figure}[htb]
  \centering
  \includegraphics[width=14cm]{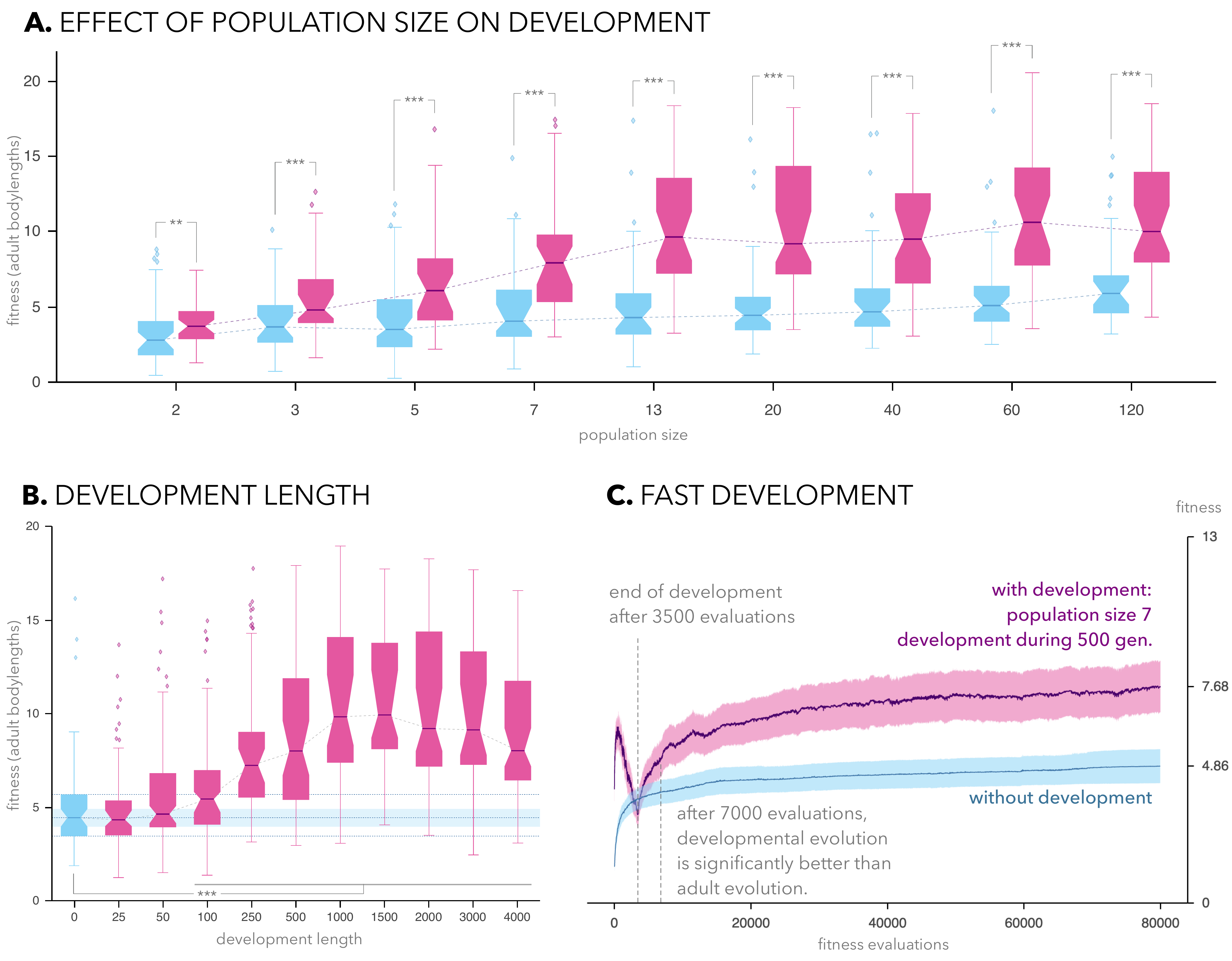}
  \captionsetup{width=14cm}
  \caption{\textbf{Development is effective a low and high population size, and \added{over}\removed{even with} short development\added{al periods}.} \textbf{A.} Comparisons of the final fitness between the adult evolution and development with starting mass 0.125, for different population size. The number of generations is kept at 4000 regardless of population size.
  \textbf{B.} Final fitness observed for different development length of the development with starting mass 0.125, from 0 (adult evolution) to 4000 (only development). The total number of generations is always 4000, with the robots spending the remaining generations evolving with the adult morphology after development ends.
  \textbf{C.}~Average fitness over 100 runs. The adult evolution (with 20 members per generation) is contrasted with development with starting mass 0.125, with a population of 7 members (2 parents, 5 children) and a development length of 500. The x-axis is the number of fitness evaluations to account for the different population sizes.}
  \label{fig:accelerated_dev}
\end{figure}

Developmental evolution is effective to increase fitness, but it requires to go through the developmental period, where the behaviors are not learned—and therefore a priori not effective—on the target adult morphology. One question is then: how fast can development be done? If we measure the length of development by the number of fitness evaluations spent on a morphology that is not the adult one, two ways to reduce it are straightforward:  reduce the size of the population or reduce the number of developmental generations.

Figure \ref{fig:accelerated_dev}.A looks at the effect of population size on the impact of development. We compare the fitness distributions of different population sizes: 2 (1 parent/1 child), 3 (1/2), 5 (2/3), 7 (2/5), 13 (3/10), 20 (5/15, the basis for all other results), 40 (10/30), 60 (15/45) and 120 (30/90). As we can observe, development brings significant improvement at every population size studied, even for the extreme case of one parent/one child. And in the range of sizes 2--13, the fitness of developmental evolution grows faster than the fitness of adult evolution as the size of the population grows. Development is both robust to small population sizes and takes advantage of population increase. This is partly explained by development not acting \emph{within} the population, like would a diversity-preserving algorithm, for instance, but \emph{on} it, uniformly.

Figure \ref{fig:accelerated_dev}.B analyzes how the length of development affects the final fitness. Even a short developmental phase, 100 generations, brings significant benefits, which then increase as development goes on longer. The slight decrease for development length 3000 and 4000 can be explained by the evolutionary search ending at generation 4000 in all cases: the settling period into the adult morphology is shortened for development length 3000 and non-existent for 4000. Development is therefore effective even when short—but not too short—, and it benefits from a long development and the presence of an extended adult phase.

When trying to reduce as much as possible the development phase, we can take advantage of the dynamics of those two dimensions above: development length and population size. After trying different combinations of population size and development lengths, a population of seven---two parents and five children---and a development length of 500 generations offers a good balance between developmental speed and fitness benefits, with a total of 3500 fitness evaluations during development. When comparing it to the adult evolution with 20 members, after 7000 fitness evaluations, i.e., 350 generations of the adult evolution, the fitness is significantly better (it is also significantly better than an adult evolution with seven members per generation, see Figure \ref{fig:accelerated_dev_7}). Interestingly, at 350 generations, the adult evolution has just started slowing down fitness improvement. So implementing development here provides little drawbacks in terms of time spent in the search.

\subsection{Comparisons with Evo-Devo Approaches}

\subsubsection*{2D Robots}

Our approach in this work is to explore a biologically implausible modification of interaction between evolution and development. One of the main difference between our approach and evo-devo approach is the relative position of the adaptation mechanism with regards to development. In our approach, development happens at the same timescale as adaption through evolutionary search. In an evo\added{-}devo approach, the development happens at a shorter timescale, during the evaluation period, with no adaptation happening. We test the impact of an evo\added{-}devo approach on our robots.

\begin{figure}[htb]
  \centering
  \includegraphics[width=14cm]{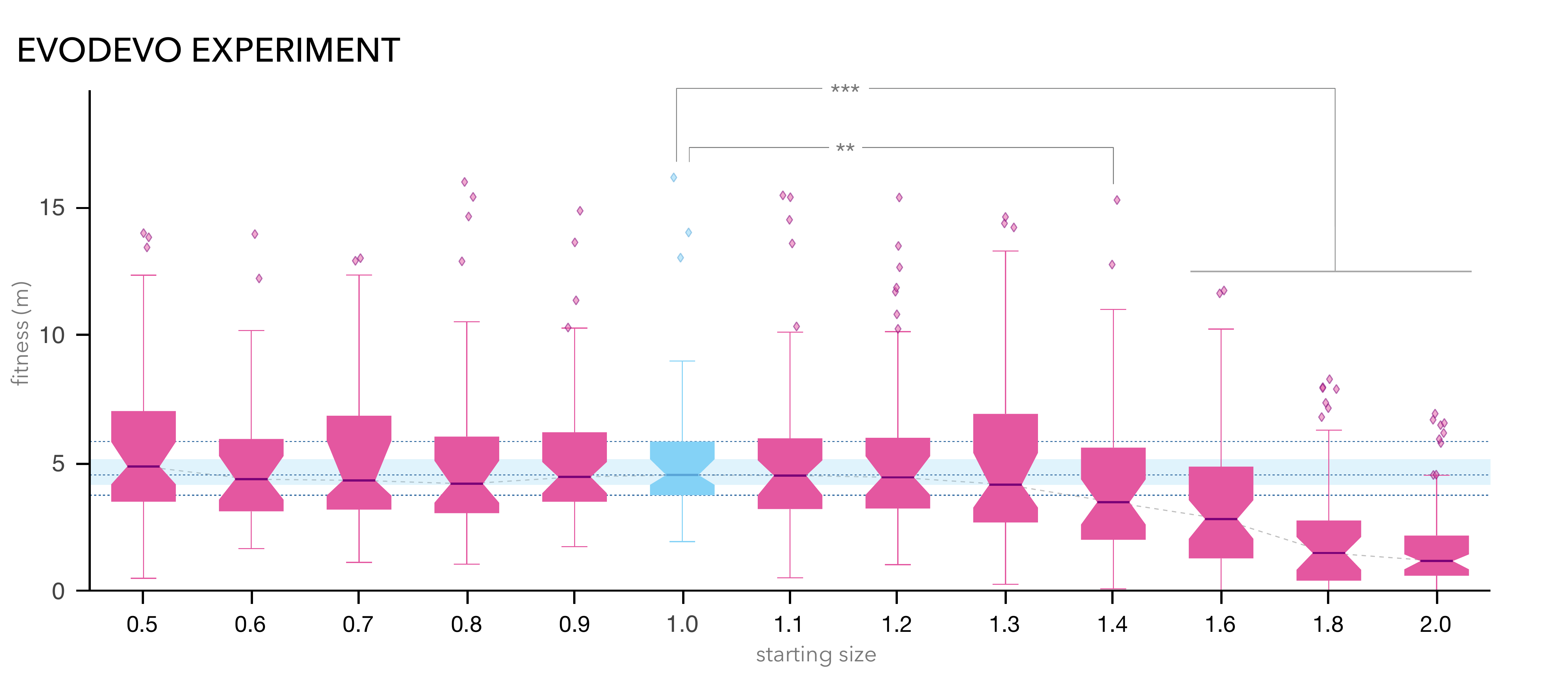}
  \captionsetup{width=14cm}
  \caption{\textbf{The evo\added{-}devo approach does not bring fitness benefits.} The boxplots represent the adult morphology performance after 4000 generations of \emph{evo\added{-}devo} evolutionary search. The blue boxplot correspond the adult evolution, equivalent to an evo\added{-}devo starting size of 1.0.}
  \label{fig:evodevo}
\end{figure}

In the evo\added{-}devo experiment, the evolutionary search still spans 4000 generations, with no developmental evolution happening. At the start of a simulation, the robots start with child body size, and settle on the ground for 9.42s. At 10s, shortly after they started to actuate, they start to grow linearly in size, to reach adult size at the 40s mark. They then continue to actuate with a fixed adult size until the 60s mark. This developmental schedule never changes during the 4000 generations. The population of behaviors of the last 50 generations of each run are then evaluated on a fixed adult morphology (adult robots during the whole 60s of simulation), and the resulting fitnesses are shown in Figure \ref{fig:evodevo}.

The evo\added{-}devo approach does not bring fitness increases over a pure adult evolution. What's more, it decreases fitness significantly for the robots that start much bigger than the adult size. This of course proves little: only that a devo\added{-}evo approach can work where a evo\added{-}devo one might not.

\subsubsection*{3D Voxel Robots}

In their 2018 article, Kriegman, Cheney, and Bongard \parencite{Kriegman2018} propose an evolutionary development approach (\emph{evo-devo}) where robots made of a 3D lattice of voxels \parencite{Hiller2014} are allowed to develop during the evaluation period. On a locomotion task, the approach outperforms a regular evolutionary algorithm (\emph{evo}), where no development happens. This task provides a good testbed to test our approach\removed{,} and \removed{to} how it generalizes to 3D environments and other robots.

For our approach—\emph{devo-evo}—we modify the \emph{evo} method by adding a mass-based developmental program on top of it. In the \emph{evo} (and \emph{evo-devo}) approach, the voxels all have a constant mass of 1 g during the simulations; in \emph{devo-evo}, we start the voxels of our robots, in the first generation, at 0.25 g, 25\% of the adult mass. The mass then increases linearly with each generation, to reach 1 g at generation 2000. The evolutionary search then continues until generation 10000. We reproduce the results of the original article (specifically, Figure 3.A) and compare them with the performance of the \emph{devo-evo} approach.

\begin{figure}[htb]
  \centering
  \includegraphics[width=10cm]{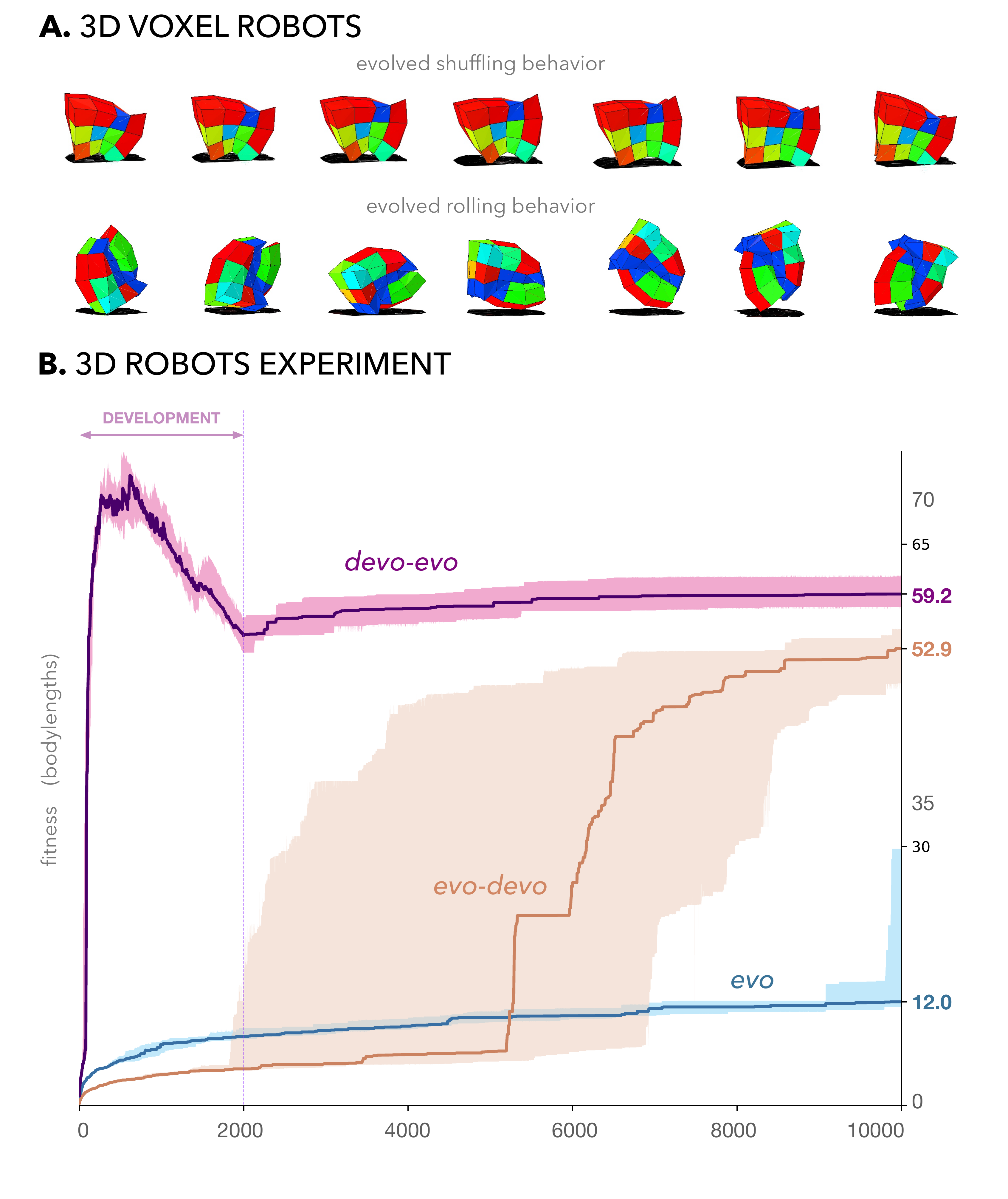}
  \captionsetup{width=12cm}
  \caption{\textbf{Devo\added{-}evo outperforms evo\added{-}devo.} \added{\textbf{A.} A 4x4x3 lattice of voxels creates robots that can move by contracting the voxels. Each voxel contracts with the same sinusoidal signal, but each has its own phase. The resting size of the voxelis specific to each voxel as well, creating robots with different morphologies. Two evolved behavior are shown: a shuffling gait at the top, and a rolling gait at the bottom. \textbf{B.}} Our \removed{evodevo}\added{devo-evo} approach shows a typical developmental performance: the performance is high when the robot is light, and then decrease as the robots become heavier during the first 2000 generations, to stabilize to high fitness that increases slowly after development ends. The graph shows the median performance of 30 runs, with the shaded area showing the 95\% boostrapped confidence interval (20000 resamplings)\added{, same as the original paper}.}
  \label{fig:evodevo_3d_median}
\end{figure}

The \removed{\emph{evo-devo}}\added{\emph{devo-evo}} approach achieves high-fitness in early development (Figure \ref{fig:evodevo_3d_median}), during the first 100 generations. The performance steadily falls as the robots' mass increases. At generation 2000, the robots have the same mass as the other approaches and the performance can be fairly compared with \emph{evo} and \emph{evo-devo}. The \emph{devo-evo} approach exhibit significantly higher median performance than either other method, and its final performance is also significantly higher than other methods. Details about the method, additional results and discussion are available in the \hyperref[sec:3drobots]{Supplementary Information}.

\section{Discussion}

Development—intrinsically—creates change. On its way to an adult form, the body goes through different iterations, each harboring the possibility to be particularly fit for a given purpose and \added{to} make some interesting behaviors easier to discover and perform. With this paper, we harness this phenomenon in robots. But, rather than evolution having only access to the global fitness of the end-result of this developmental process, we make all the developmental steps directly accessible to evolution's selection capabilities.

This is illustrated nicely in this article by how difficult rolling behaviors can be to discover for 2D and 3D robots alike, and how easier a developing morphology create an easy access to them. Kriegman, Cheney, and Bongard noticed a similar effect when adding development during fitness evaluation \parencite{Kriegman2018}. Because in many real-world problems interesting behaviors represent a very small portion of the search space, sometimes without good gradients to help guiding toward them, methods to help discovering them are crucial for robots facing the complexity of the real world. In our work, we observed that once discovered through development, those high-fitness behaviors have a chance to be maintained by evolution into the adult form. Development is \added{used} here, quite straightforwardly, \removed{used} as a source of behavioral diversity evolution can select from.

Many approaches in evolutionary robotics have aimed at fostering diversity. Some place behavioral novelty directly under evolutionary control, using selective pressures \parencite{Risi2009, Doncieux2014}, while others, like illumination algorithms \parencite{Cully2015, Mouret2015}, modify the structure of the search to expose an ensemble of solutions across specific dimensions of behavioral diversity. Our devo-evo approach does not employ either of those mechanisms: rather, morphological change is added on top of a classic evolutionary search, with no change to the evolutionary algorithm or the evaluation budget. The approach does not need memory archives to store elites or estimate novelty or need to define novelty distance or behavioral dimensions along which a grid of MAP-Elites cells should be created. Rather, it relies on pure embodiment. It is worth considering here an extreme case: even without adaptation capabilities, a robot executing the same motor activation and undergoing morphological development would produce behavioral diversity as its body grows (see Fig. \ref{fig:size}.A for an example). Irrespective of the controller of the robot, morphological change produces behavioral change.

Morphological development also creates a constant source of perturbation, an imbalance in the evolutionary search that straightforwardly deals with two fundamental issues of evolutionary computation: the bootstrap problem and premature convergence \parencite{Hornby2006, Mouret2009, Schmidt2010, Bongard2010, Eiben2015}. Premature convergence is prevented because the early \removed{constant}\added{ongoing} morphological changes modify the fitness landscape with every generation and prevent the evolutionary search from settling too early: you cannot stay in a fitness landscape's valley if the valley disappears under you. The bootstrap problem—\added{which happens} when all the members of the initial population obtain minimal fitness, and hence stall the search—is mitigated by developmental evolution, since it will, throughout development, evaluate the population \added{on} a succession of slightly different fitness landscapes. If one of those fitness landscapes creates interesting behaviors, the evolutionary search can start. In this work, we have seen that even when the robots start with eight times the adult mass, and therefore can hardly move from the starting position (Fig. \ref{fig:baselines}.C)—a good setup for a bootstrapping problem—, the fitness performance at the end of the evolutionary search is not statistically distinguishable from the one of the adult evolution (Fig. \ref{fig:sizestiffmass_boxplots}.C). That developmental evolution was resilient \added{to} all the developmental paths we explored was an unexpected observation.

Another surprising result was that evolution was able to benefit from developmental paths that made the task intuitively harder (e.g., a smaller starting size). Making a robot stronger or lighter is often an effective way to boost performance. So it came without much surprise that the developmental paths that started that way (high muscle strength or low starting mass), and hence made the task initially easier, led to increased adult performance: this is a typical incremental evolutionary strategy. The results of the lower starting size experiments are harder to explain. It is indeed tempting to intuitively conclude that physical immaturity is a hindrance to skill acquisition only made necessary in \added{animals}\removed{biology} by physiological constraints. This line of thinking, whether implicit or explicit, combined with considerations of practicality, robustness and feasibility, has been embraced by robotics: robots are born adult. This study, however modest its results, suggests that robots that grow up in size can bring qualitative performance benefits.

And the reason may lie in how developmental evolution affects exploration. The creation of behavioral diversity and the changing fitness landscape preventing early convergence both respectively facilitate and stimulate exploration. In the analysis, we have shown that increased exploration (characterized by increased genotyped distance between the first and last population) correlates positively with fitness within a developmental condition. The lower starting size experiments show this in another way: the smaller you start, the higher the adult performance. Starting smaller means a faster pace of change during development, and more morphologies explored.

\added{Our study reaches a different conclusion than \cite{Bongard2011} paper. There's a lot of differences---robot (rigid), task (phototaxis), fitness criterion (time to success), many aspects of the method---that make a straightforward comparison difficult. One issue that Bongard reported was abrupt body changes (the evolutionary process goes through four discrete morphologies, rather than developing gradually), leading to a drop of fitness, in some case to zero. This may explain why performance was affected negatively. Our developmental paths change morphology slowly, and this may make them robust to this. Future work is needed to investigate those issues in detail.}

We have explored only a few developmental programs in this article. And, much the same that humans are not necessarily good at hand-tuning good controllers for complex body shapes, designing developmental programs, as they become more complex and dynamic, might be beyond the skills of human designers. We may then let developmental programs evolve themselves, which may raise issues in how expensive in computing resources that might be. Nevertheless, the commonalities seen in the biological world in the development of species with widely different morphologies teases the possible discovery of general developmental principles and mechanisms that could be effectively applied to a wide range of robots.

As well as adapting to different robots, a developmental program would be useful if it can benefit a large range of tasks. One fundamental motivation for our approach is the idea that a well-designed development program for robots might adapt well to as many different tasks and environments, \removed{because that is what happens in biology}\added{much like what we observe in humans}.

\section{Conclusion}

Developmental evolution gradually changes the morphology of populations of robots during an evolutionary search. In this study, we have seen that spending evolutionary time on morphologies different than the target one, potentially smaller or weaker ones, may bring significant improvement to fitness. This is achieved in  two ways: by making some efficient behaviors easier to discover, and by preventing premature convergence through a continuous modification of the search space, therefore fostering its exploration. Because morphological development is intrinsic to the robot, \removed{it}\added{its implementation} does not depend on the task, environment, or fitness function. It has shown to be robust to many different developmental paths, to work with small and large population sizes, on short and long developmental lengths, and with 2D and 3D robots. Interesting parallels exist between developmental evolution and developmental learning, and inspirations and techniques from both could lead to fruitful future work.

\printbibliography
\newpage

\pagebreak
\newpage

\beginsupplement

\section*{Supplementary Information}
\label{sec:si}

The source code for the experiments is available at \url{https://doi.org/10.6084/m9.figshare.14879988}. \added{You can contact Fabien Benureau at fabien.benureau@oist.jp (permanent email: fabien@benureau.com) for questions about the code and data.}

\subsection*{Movies}
\label{sec:movies}

Movies S1, S2, S3 are available at \url{https://doi.org/10.6084/m9.figshare.13147790}.

\subsection*{Simulation Details}
\label{sec:sim_details}

We use an impulse-based physics engine to simulate the robots. The timestep is set at 0.005 seconds. Each point mass experiences friction forces $f_f$ as well as reaction forces $f_r$ when in contact with the ground, and the forces of all springs connected to it $\sum_{i}^{\textrm{links}} f_{l_i}$ and gravity $g$:
\[m \ddot{\mathbf{p}} = f_r(\dot{\mathbf{p}}) + f_f(\dot{\mathbf{p}}) + g + \sum_{i}^{\textrm{links}} f_{l_i} \]
With $p$ the position of the point mass.

Contact with the ground generate reaction force $f_r(\dot{\mathbf{x}})$; the floor is flat and has a restitution of 0.2. For the friction force $f_f(\dot{\mathbf{x}})$, all nodes have a static friction coefficient of 0.5, except for the ones of the tip of each tentacle whose friction is 1.0. We implement dynamic friction: when a node is moving along the ground at more than 1 mm/s, the dynamic friction is half of the static friction. The force of gravitation $g$ is arbitrarily set at -100 m/s$^2$ in the direction of the y-axis. The forces $f_l$ of the springs are implemented using soft-constraints \parencite{Catto2011}, in a similar way \added{that} they are implemented in the ODE simulator, giving us springs simulated with first-order implicit integration, hence enhancing numerical stability.

To help with numerical stability, all the values of the characteristics of the robots were scaled up. Size 1.0 corresponds to a section height of 4 m. The width of all sections is 3 m. The adult stiffness of the muscles is 20$\,$000 N/m. The adult mass of the nodes is 1 kg, and the gravity is -100 m/s$^2$. The damping ratio of all springs is always~1. The rigid diagonal links of each section, as well as the two tips springs and the springs of the center of the robot have stiffness 500$\,$000~N/m. The passive flexible springs at the top and bottom of each section have 65$\,$000~N/m stiffness (except the first section, where the springs shared with the center is rigid at 500$\,$000~N/m). The stiffness values were chosen so the adult morphology could move around yet still \added{be} slightly underpowered so that both increase and decrease in stiffness would significantly change behavior.

\subsection*{Genotype and Mutations}

The genotype of each robot is a 24-scalar vector, with each tentacle corresponding to 4 scalars. The motor commands of each of the two motor groups of a tentacle are encoded using two scalars, and each motor group has an independent phase in $[-\pi, \pi]$ and amplitude in $[0, 0.2]$. Initial random genotypes are generated by choosing values uniformly in those ranges. Each parent creates offspring by mutating two values of the 24-scalar vector. Given a normalized value in [0, 1], the mutation is a random normal perturbation with variance 0.05. When above or below the boundaries, the excess is mirrored back within the range.

\subsection*{Motor Noise}

Motor noise is introduced at each timestep of the simulation, as an independent normal perturbation on each muscle with variance $10^{-4}$.

\subsection*{Experiments with 3D Voxel Robots}
\label{sec:3drobots}

\subsubsubsection{\emph{Method}}

Kriegman, Cheney, and Bongard consider, in their 2018 article, an evolutionary search that includes development during the evaluation period. Robots made of voxels (see Fig \ref{fig:evodevo_3d_median}) possess a genotype that encodes the size of each voxel, as well as the phase of the contraction of the voxels. In their \emph{evo-devo} approach, the robots change morphology as they behave during simulation. Each voxel has two sizes, a start and an end size. A voxel begins the simulation with the start size and grows (or shrinks) linearly toward the end size until the simulation finishes. The development is under evolutionary control: those two sizes are encoded in the genotype of the robots. Similarly, the phase of the sinusoïdal contractions of the voxel develops: during the simulation, the phase shift from a start phase to an end phase, similarly encoded in the genotype (for specific details of the \emph{evo} and \emph{evo-devo} methods, we refer the reader to the original article \parencite{Kriegman2018}).

Our \emph{devo-evo} approach adds a mass-based developmental program on the top of the regular \emph{evo} algorithm, which uses only one size and one phase per voxel. We chose the mass-based developmental program rather than a size-based one because changing the size of a cube voxel is isomorphic, and therefore unlikely to create different morphological dynamics for our robots. Changing the mass was also one of the easiest change to implement in the simulation, and was orthogonal to the evolutionary search on the size of the voxels. Compared to \emph{evo-devo}, we do not develop the controller of the robot.

The experimental parameters were kept identical to the original article, and the published source code by the authors was used to reproduce their results (specifically their Figure 3.A). We do not obtain the exact same results as the original article, but our reproduced results \added{are}\removed{appear fairly} consistent with the published ones. The source code with the modifications for the \emph{devo-evo} approach is available at \url{https://doi.org/10.6084/m9.figshare.14879988}.

The simulations for \emph{evo} and \emph{evo-devo} are made with zero actuation noise. This allows the surviving parents not to be reevaluated in a new generation, as the fitness they obtain would be identical to the one computed in the previous simulation. In the \emph{devo-evo} approach, with morphological change creating different conditions for every generation of the developmental period, we cannot rely on such a mechanism. Instead, we reevaluate the surviving parents during the 2000 generations of the developmental period, doubling its evaluation cost. \added{To account for this extra cost of the \emph{devo-evo} approach, }\removed{At similar evaluation cost,} the \emph{devo-evo} performance at generation 2000 and 8000 should be compared with the \emph{evo-devo} performance at generation 4000 and 10000 respectively. This increased cost for the \emph{devo-evo} method is present when parents are surviving and not reevaluated. In Figure \ref{fig:evodevo_kriegman_pop} we show that this is actually not an issue, as \emph{devo-evo} can obtain significantly the same performance on a population half the size.

\subsubsubsection{\emph{Results and Discussion}}

In Figure \ref{fig:evodevo_3d_median}, our approach, \emph{devo-evo}\added{,} achieves high-fitness \removed{right from} early \added{in} development. While, during development, with the lower mass of the robots, it does not make sense to compare the fitness of the different approaches, at the end of development, at generation 2000, the task is the same for all approaches—the robots all have the same mass. At that point, the \emph{devo-evo} method has discovered high-fitness behaviors significantly faster than \emph{evo-devo} at generation 4000. The final performance at generation 8000 is also significantly higher than the one of the \emph{evo-devo} approach at 10000.

\begin{figure}[htb]
  \centering
  \includegraphics[width=12cm]{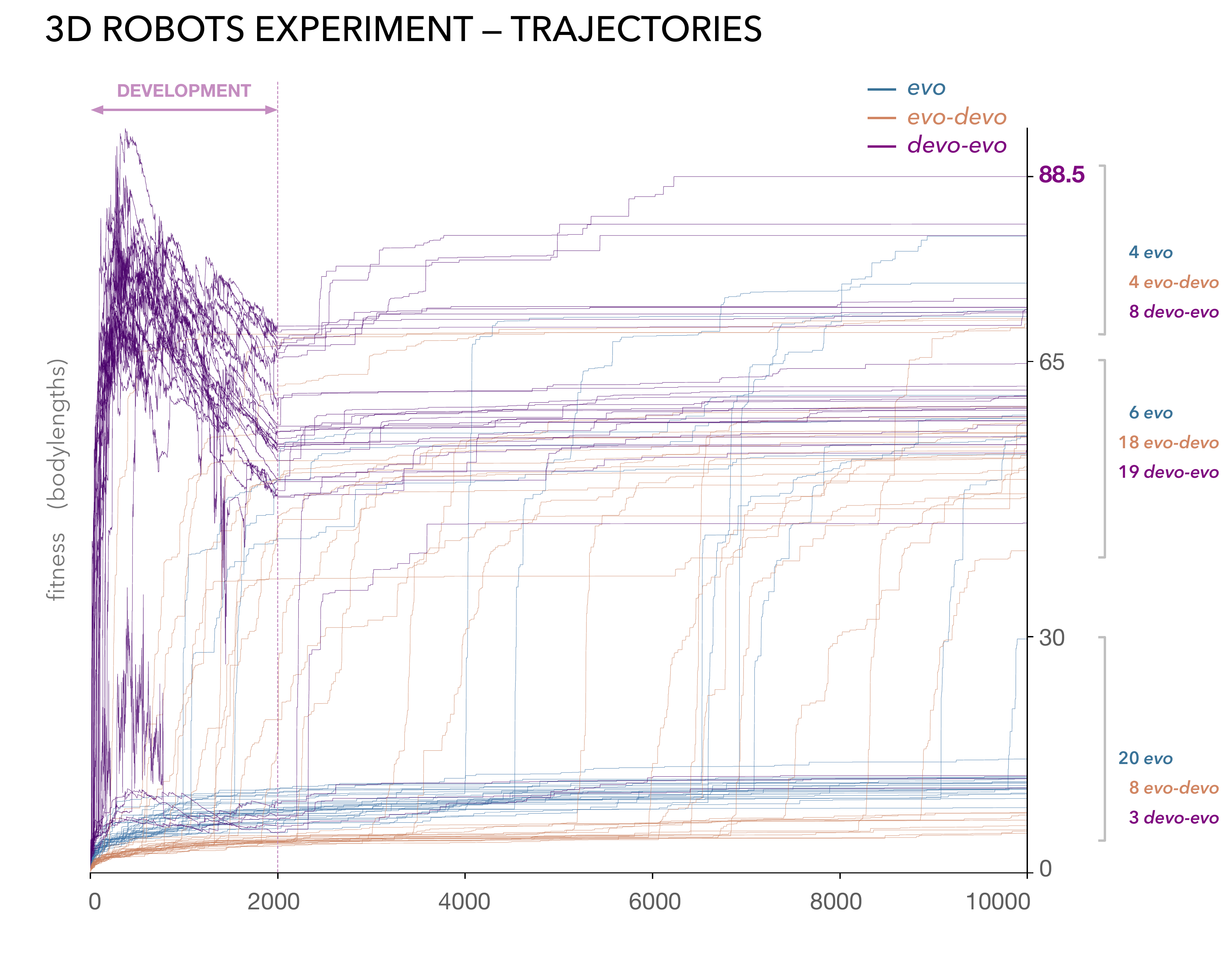}
  \captionsetup{width=12cm}
  \caption{\textbf{} Fitness trajectories of each of the 30 runs of the \emph{evo}, \emph{evo-devo}, and \emph{devo-evo} methods. Each curve is the fitness of the champion of each generation of a run. We can see the developmental patterns of the \emph{devo-evo} approach during the \added{first} 2000 generations. Some runs of \emph{devo-evo} do not discover rolling behavior (fitness < 30 bodylengths), but achieve performance competitive with the shuffling \emph{evo} behaviors. Two classes of rolling behaviors are exhibited by the search, which we separate with a 65-fitness cut-off point (see also Figure \ref{fig:evodevo_kriegman_mass}.D for an even clearer illustration of the pattern). On the right of the graph, the number of runs per approach is depicted for shuffling (fitness < 30), low-fitness rolling (30 < fitness < 65) and high-fitness rolling (65 < fitness).
  }
  \label{fig:evodevo_3d_traj}
\end{figure}

To better understand the dynamics of the evolutionary search for each approach, we show all the evolutionary trajectories of the runs in Figure \ref{fig:evodevo_3d_traj}. We observe a branching behavior, with the trajectories shooting upward from low-fitness to high-fitness when discovering rolling behaviors.

\begin{figure}[htb]
  \centering
  \includegraphics[width=13cm]{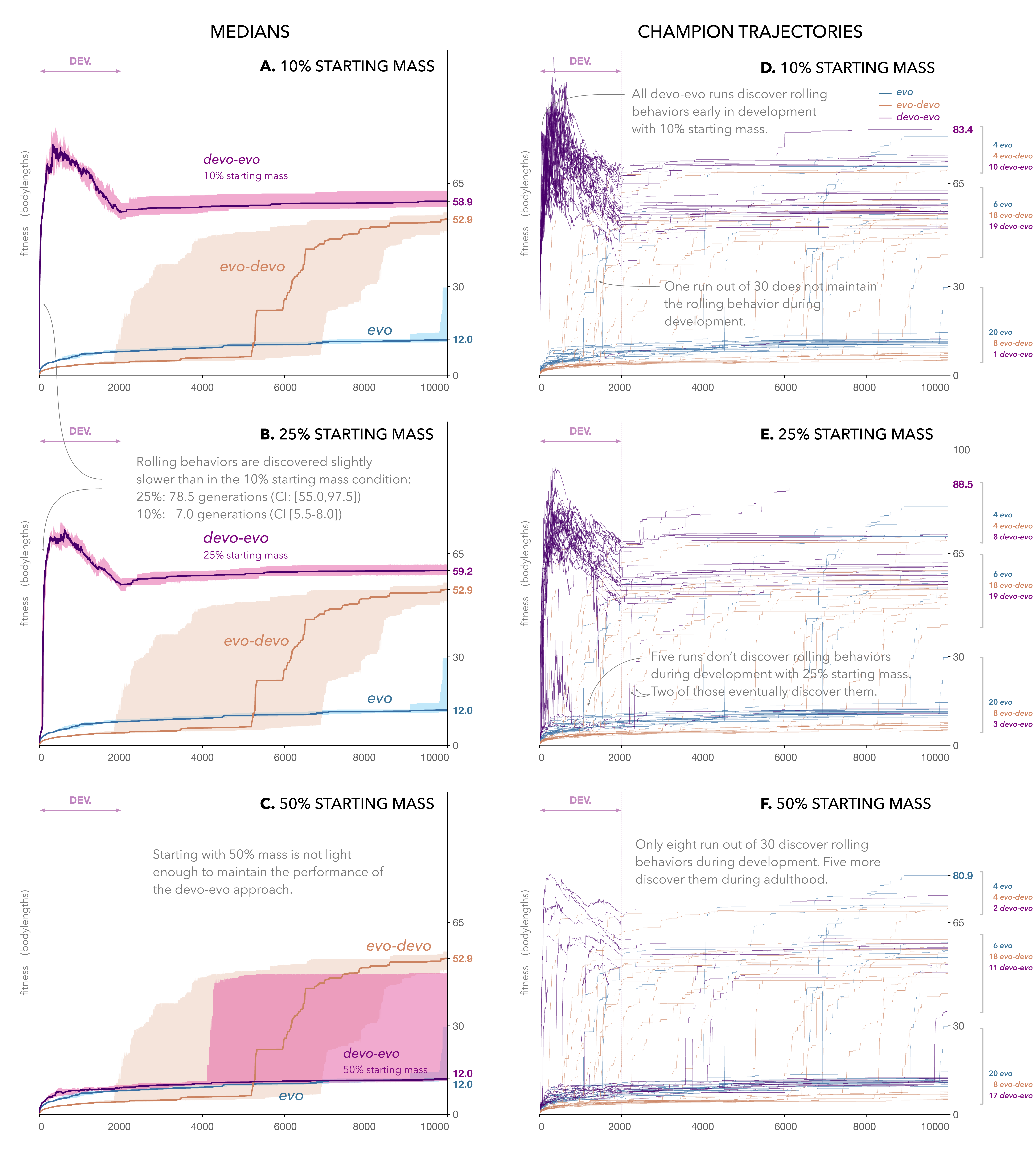}
  \captionsetup{width=13cm}
  \caption{\textbf{Starting with less mass ensure discovery of rolling behaviors.} The lower the starting mass, the higher the percentage of runs discovering high-performing behaviors. \textbf{A}, \textbf{B}, and \textbf{C} show the median of each approach with the 95\% bootstrapped condidence interval (20000 bootstrap samples), and \textbf{D}, \textbf{E}, and \textbf{F} show the trajectories of the champion for each generation of each run. In \textbf{B}, the median discovery time is calculated among the runs that discovered rolling behavior, and the first generation achieving a fitness of more than 30 is taken into account. The confidence interval is bootstrapped with 20000 bootstrap samples.
 }
  \label{fig:evodevo_kriegman_mass}
\end{figure}

\begin{figure}[htb]
  \centering
  \includegraphics[width=13cm]{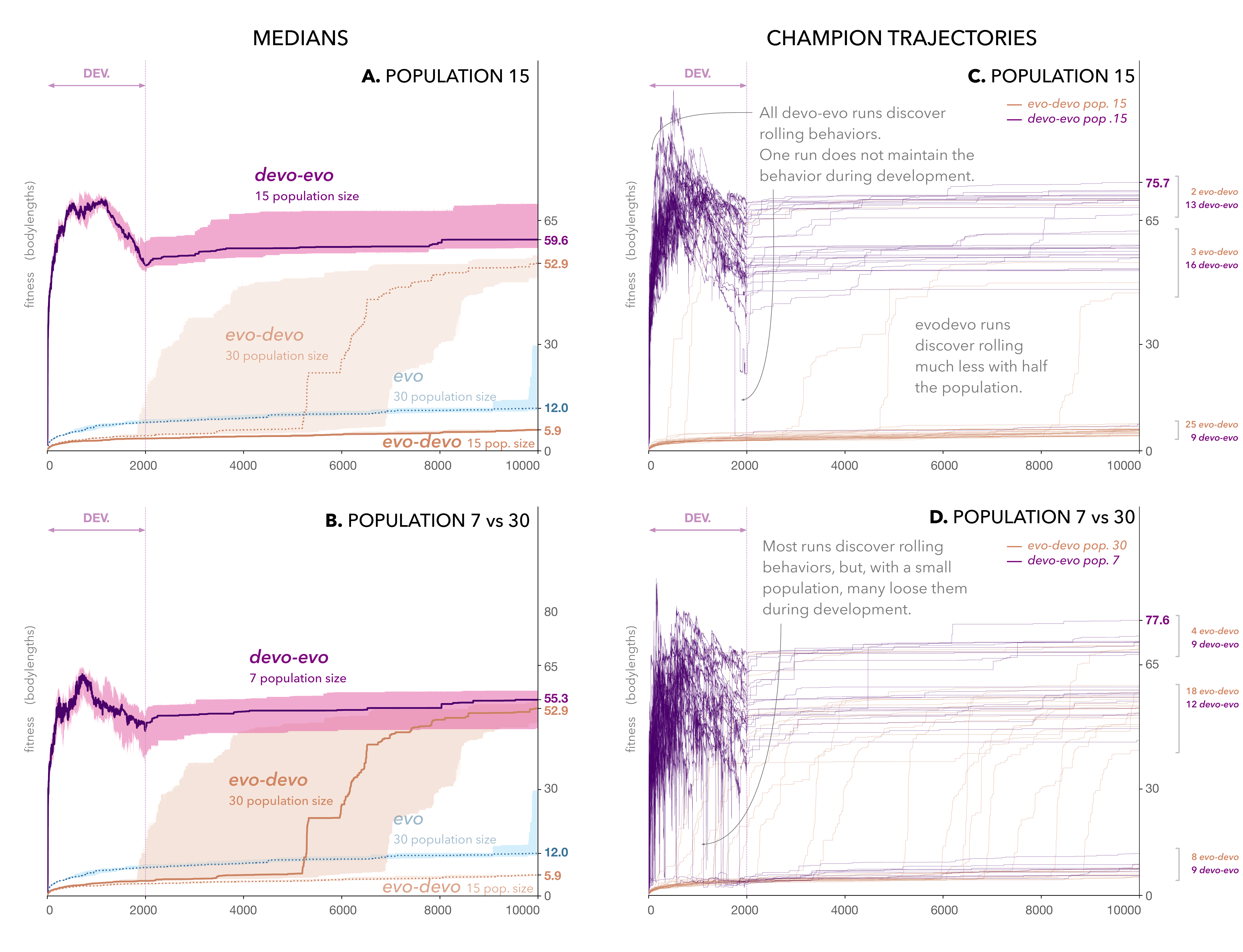}
  \captionsetup{width=13cm}
  \caption{\textbf{Devo-evo is robust to small population sizes.} We compare the performance of the \emph{devo-evo} and \emph{evo-devo} approach when we reduce the population size. \textbf{A.} and \textbf{C.} With half the population size (15), the \emph{devo-evo} method conserves its performance, while the \emph{evo-devo} does not discover rolling behaviors anymore. \textbf{B.} and \textbf{D.} In an more extreme comparison, we compare the \emph{devo-evo} approach at one-fourth of the population (7) with the \emph{evo-devo} at the original population size \added{(30)}: their performance are not significantly different. While the \emph{devo-evo} approach requires the parents to be reevaluated during the developmental period, its robustness to a small population size actually makes it much less expensive in evaluation budget. Note: for the voxel 3D robots experiments, we follow the original article conventions: the population size correpsond to the numbers of children at each generation. In the rest of the article, we have included the parents in that count.
 }
  \label{fig:evodevo_kriegman_pop}
\end{figure}

\begin{figure}[htbp!]
  \centering
  \includegraphics[width=12cm]{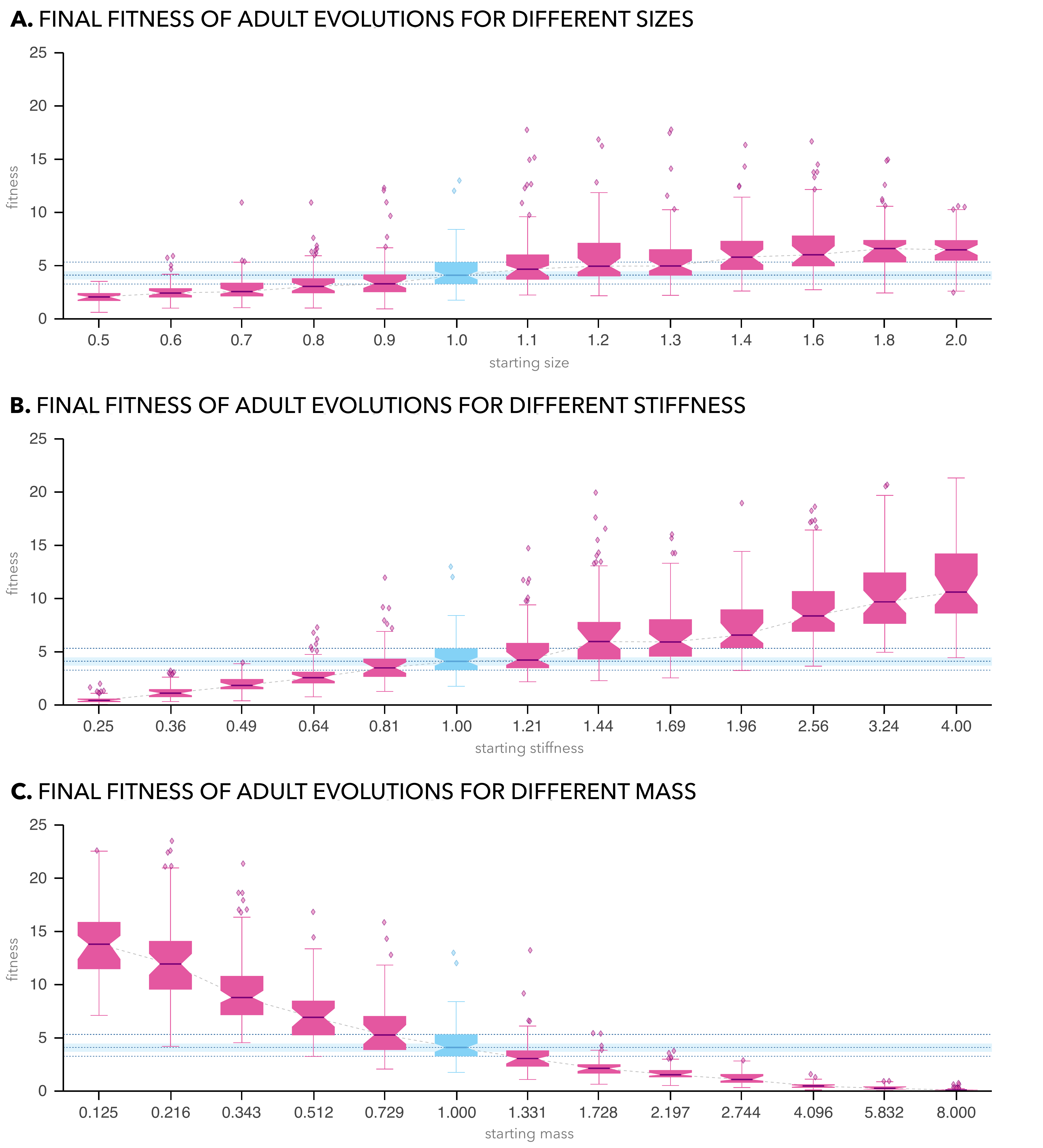}
  \captionsetup{width=10cm}
  \caption{Same as Figure \ref{fig:dev_muscle_results}.D, but for the results of A, B, and C of Figure \ref{fig:sizestiffmass_boxplots}, respectively.}
  \label{fig:baselines}
\end{figure}

\begin{figure}[htbp!]
  \centering
  \includegraphics[width=12cm]{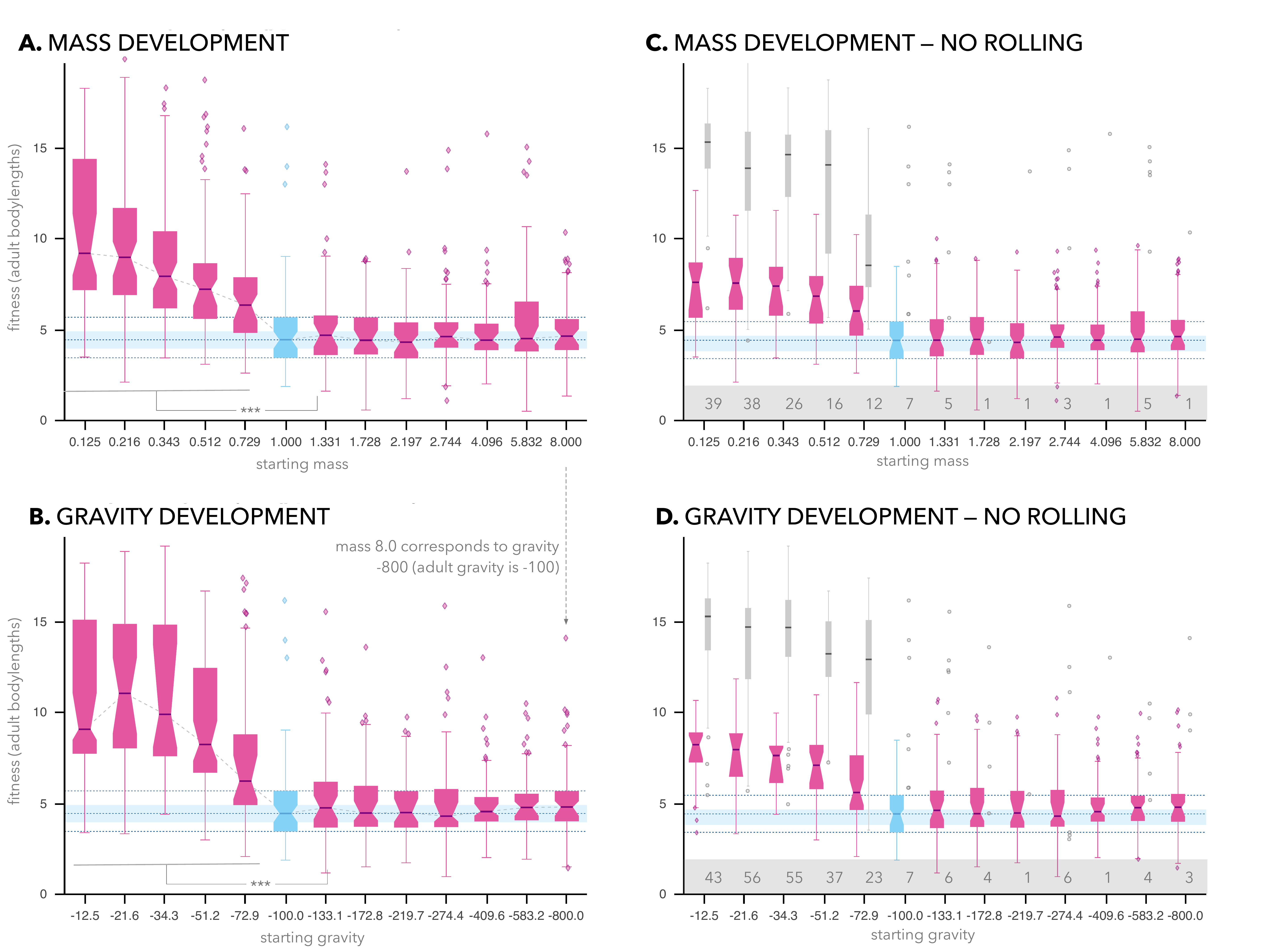}
  \captionsetup{width=11cm}
  \caption{\textbf{Developing gravity gives similar results as developing mass}. \textbf{A.} and \textbf{C.} Figure \ref{fig:sizestiffmass_boxplots}.C and Figure \ref{fig:sizestiffmass_boxplots}.F, reproduced here for convenience. \textbf{B.} \added{and \textbf{D.}} Results for the development of gravity. Gravity develops cubically, in the same fashion as mass. The boxplots are aligned vertically, the reference gravity is -100 \added{m/s$^2$} for mass 1.0, so for instance mass \added{0.125} is gravity \added{-12.5}. The results of gravity are similar to mass, with higher medians.}
  \label{fig:s_gravity}
\end{figure}

\begin{figure}[htbp!]
  \centering
  \includegraphics[width=8cm]{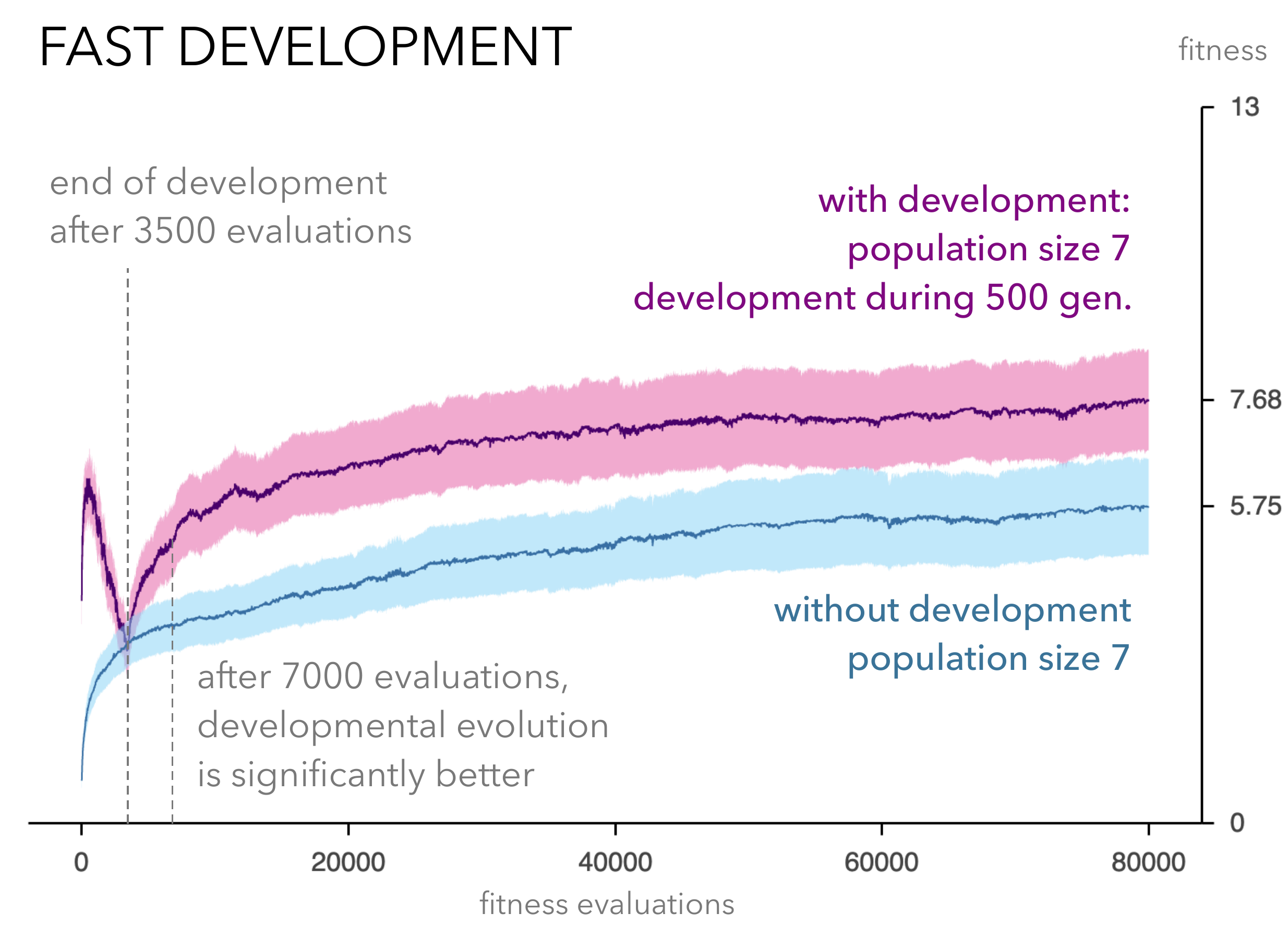}
  \captionsetup{width=10cm}
  \caption{Same as Figure \ref{fig:accelerated_dev}.C, but the adult evolution operates with a population size of 7 (2 parents, 5 children), and hence, 11428 generations. This high number of generations explain the final fitness being slightly higher than the final fitness of the adult evolution of Figure \ref{fig:accelerated_dev}.C. \removed{Let's also note that in Figure \ref{fig:accelerated_dev}.A, the number of generations is kept at 4000, hence the lower fitness distribution that displayed here.}}
  \label{fig:accelerated_dev_7}
\end{figure}

\begin{figure}[htbp!]
  \centering
  \includegraphics[width=14cm]{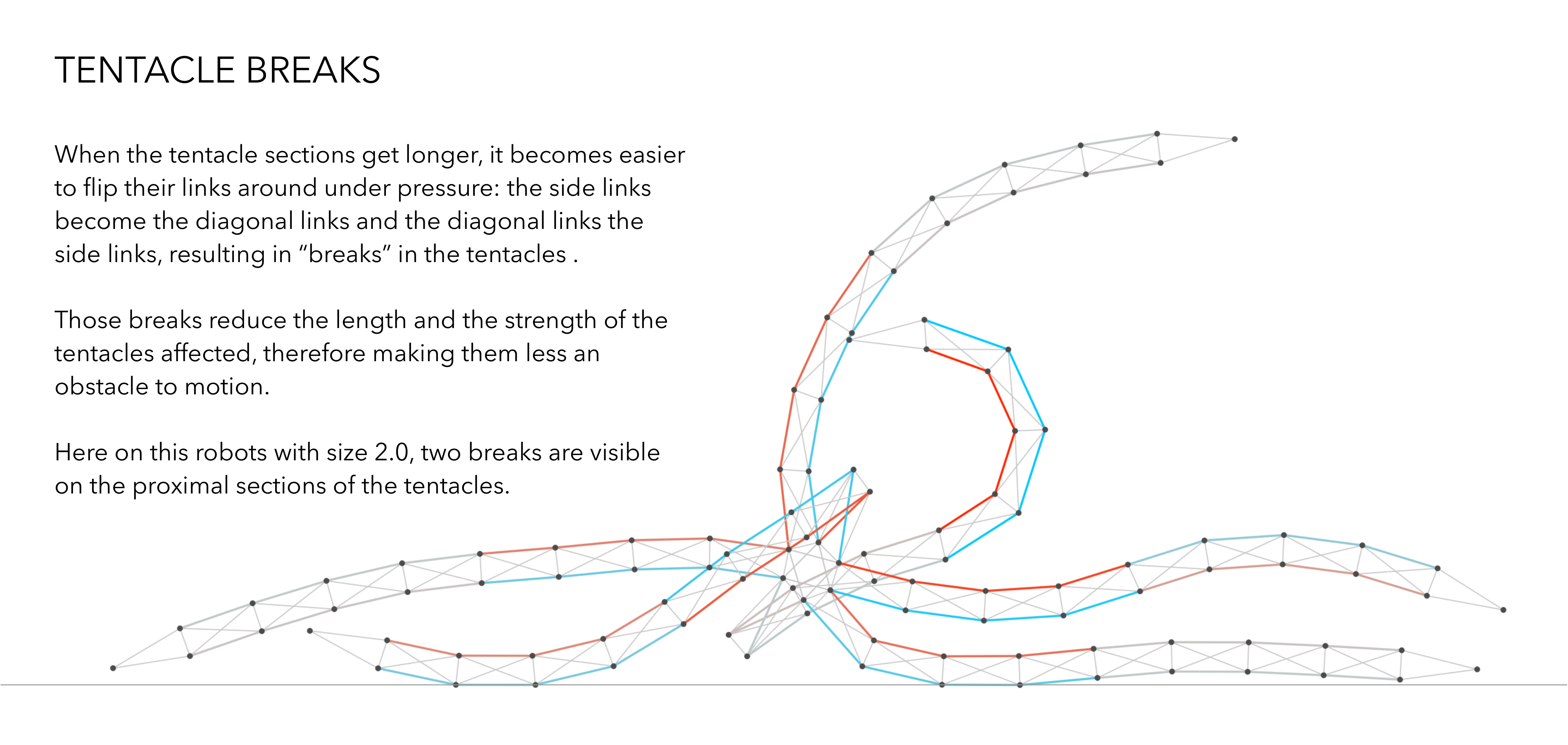}
  \captionsetup{width=10cm}
  \caption{An example of tentacle breaks.}
  \label{fig:tentacle_breaks}
\end{figure}

In Figure \ref{fig:evodevo_kriegman_mass} we can see the impact of the starting mass. With low starting mass (10\% of the adult mass), all runs discover the rolling behaviors in early development. One run, however, does not maintain the behavior into adulthood. With 25\% starting mass—the results presented \removed{above} \added{in Figures \ref{fig:evodevo_3d_median} and \ref{fig:evodevo_3d_traj}}—five runs do not discover the rolling behavior, although two of those will stumble on them during adulthood. If starting with only 50\% of the adult mass, most runs do not discover rolling during development (only eight do, with five more during adulthood).

It is interesting to notice the timing of the discovery of rolling during development across the three mass conditions. For 10\% starting mass, the discovery is made within the first few generations for all runs (median is 7.0 generations with CI [5.5-8.0]), while for 25\% starting mass the discovery is more spread out (median is 78.5 generations with CI [55.0,97.5]). This is even more pronounced for the 50\% starting mass condition, where the discovery happens half of the time during the first thousand generations for different runs (median is 1033.0 generations with CI [99.0, 3752.0], all medians are computed within the runs that discover rolling behaviors. For the \emph{evo-devo} approach, the median discovery time is 2569.0 generations with CI [1555.0, 5294.0], while for the \emph{evo} it is 5538.0 generations with CI [2676.0, 6902.0]). In conclusion, the 10\% starting mass creates trivial access to those rolling behaviors, and the discovery becomes progressively harder as mass increases.

Another point worth mentioning is that the performance of the \emph{devo-evo} behaviors that do not discover or lose rolling, across all the mass conditions, remains competitive with the non-rolling \emph{evo} performance, thereby illustrating the robustness of \emph{devo-evo} when the developmental program does not bring benefits.

Another illustration of the robustness of \emph{devo-evo} is with regard of population size, as we already shown \added{in} Figure \ref{fig:accelerated_dev}.A. In Figure \ref{fig:evodevo_kriegman_pop}, we show that the \emph{devo-evo} method still performs well when we divide the population size by 2 or 4. This also counterbalances completely the additional cost of needing to reevaluate the parents during the developmental period, and make \emph{devo-evo} cheaper in the number of evaluations.

In conclusion, we showed that \emph{devo-evo} is robust and effective on another robotic setup from the literature, and that it allows qualitative jumps in performance without introducing much complexity into existing evolutionary algorithms.

\newpage
\pagebreak
\subsection*{Additional Figures}

\end{document}